%% file: neurips_2025.tex
\documentclass{article}

\PassOptionsToPackage{square,sort&compress,numbers}{natbib}

\usepackage[final]{neurips_2025}

\usepackage[utf8]{inputenc} %
\usepackage[T1]{fontenc}    %
\usepackage{hyperref}       %
\usepackage{url}            %
\usepackage{booktabs}       %
\usepackage{amsfonts}       %
\usepackage{nicefrac}       %
\usepackage{microtype}      %
\usepackage{xcolor}         %
\usepackage{graphicx}
\usepackage{amsmath}
\usepackage{bm}
\usepackage{bbm}
\usepackage{mathtools}
\usepackage[inline]{enumitem}
\usepackage{mdframed}
\usepackage{xcolor}
\usepackage{minitoc}
\usepackage{algorithm}
\usepackage{algpseudocode}

\usepackage{tikz}
\usetikzlibrary{arrows.meta,positioning}

\usepackage{amsthm}

\theoremstyle{plain}
\newtheorem{theorem}{Theorem}[section]  %
\newtheorem{lemma}[theorem]{Lemma}

\theoremstyle{definition}
\newtheorem{definition}[theorem]{Definition}

\newtheoremstyle{noparens}
  {\topsep}   %
  {\topsep}   %
  {\itshape}  %
  {}          %
  {\bfseries} %
  {.}         %
  {.5em}      %
  {\thmname{#1}\thmnumber{ #2}\thmnote{ #3}} %
\theoremstyle{noparens}
\newtheorem*{theorem*}{Theorem} 

\makeatletter
\pretocmd{\@thm}{\vspace{\dimexpr\topsep\relax}}{}{}
\makeatother

\DeclareMathOperator*{\argmin}{arg\,min}

\title{Scaling can lead to compositional generalization}

\author{%
  Florian Redhardt\thanks{Equal contribution}\\[0.5ex]
  ETH Zurich\\
  \And  
  Yassir Akram\footnotemark[1]\\[0.5ex]
  ETH Zurich\\
  \And
  Simon Schug\thanks{Correspondence to \texttt{sschug@princeton.edu}.} \\ [0.5ex]
  Princeton University \\
}

\begin{document}
\doparttoc %
\faketableofcontents %
\maketitle
\setcounter{footnote}{0}
\input{abstract}
\input{main}
\newpage
\bibliographystyle{unsrtnat}
\bibliography{bibliography}
\newpage
\renewcommand \partname{}  %
\renewcommand \thepart{Appendix}  %
\addcontentsline{toc}{section}{}  %
\part{}  %
\parttoc %
\include{appendix}
\end{document}

%% file: abstract.tex
\begin{abstract}
Can neural networks systematically capture discrete, compositional task structure despite their continuous, distributed nature?
The impressive capabilities of large-scale neural networks suggest that the answer to this question is yes.
However, even for the most capable models, there are still frequent failure cases that raise doubts about their compositionality.
Here, we seek to understand what it takes for a standard neural network to generalize over tasks that share compositional structure.
We find that simply scaling data and model size leads to compositional generalization.
We show that this holds across different task encodings as long as the training distribution sufficiently covers the task space.
In line with this finding, we prove that standard multilayer perceptrons can approximate a general class of compositional task families to arbitrary precision using only a linear number of neurons with respect to the number of task modules.
Finally, we uncover that if networks successfully compositionally generalize, the constituents of a task can be linearly decoded from their hidden activations.
We show that this metric correlates with failures of text-to-image generation models to compose known concepts.
\\[1ex] \footnotesize{Code available at \url{https://github.com/smonsays/scale-compositionality}}
\end{abstract}

%% file: main.tex
\section{Introduction}
The ability to understand and produce novel combinations from familiar constituents is a key faculty of intelligence.
It has been debated for decades whether neural networks are ever able to truly achieve such compositional generalization \citep{fodor_connectionism_1988}.
Regardless of these theoretical considerations, scaling neural networks continues to result in increasingly capable models \citep{kaplan_scaling_2020, brown_language_2020, wei_emergent_2022}.
Naturally, as models are scaled up, their capacity to memorize grows, and it is perhaps unsurprising that as a result of training on ever larger datasets their ability to recall more information grows too \citep{mahajan_exploring_2018}.
However, the nature of compositionality is an exponential growth and ultimately any attempt to exhaustively capture this breadth by scaling the training data will be confronted with physical constraints.

Many works therefore advocate that neural network architectures should be explicitly endowed with compositional structure \citep[e.g.,][]{boopathy_breaking_2025,battaglia_relational_2018, russin_systematicity_2020, du_compositional_2024} to allow making \textit{infinite use of their finite means} \citep{von_humboldt_humboldtlanguage_1836, chomsky_aspects_1965}.
Capturing the underlying compositional procedure of the data is a more efficient pathway to generalize.
In particular, the algorithmic complexity of this generalizing solution is much smaller than the complexity of the memorizing solution \citep{ren_understanding_2024}.
But does this mean that architectures need to explicitly factorize according to the data's underlying compositional mechanisms
\citep{du_compositional_2024}?
For instance, monolithic networks have been shown to discover modular subnetworks which may enable compositionality without specialized symbolic mechanisms \citep{lepori_break_2023}.
Maybe simply scaling the data and size of neural networks is then enough to achieve compositionality.
\begin{figure}[t]
    \centering
    \includegraphics[width=\textwidth]{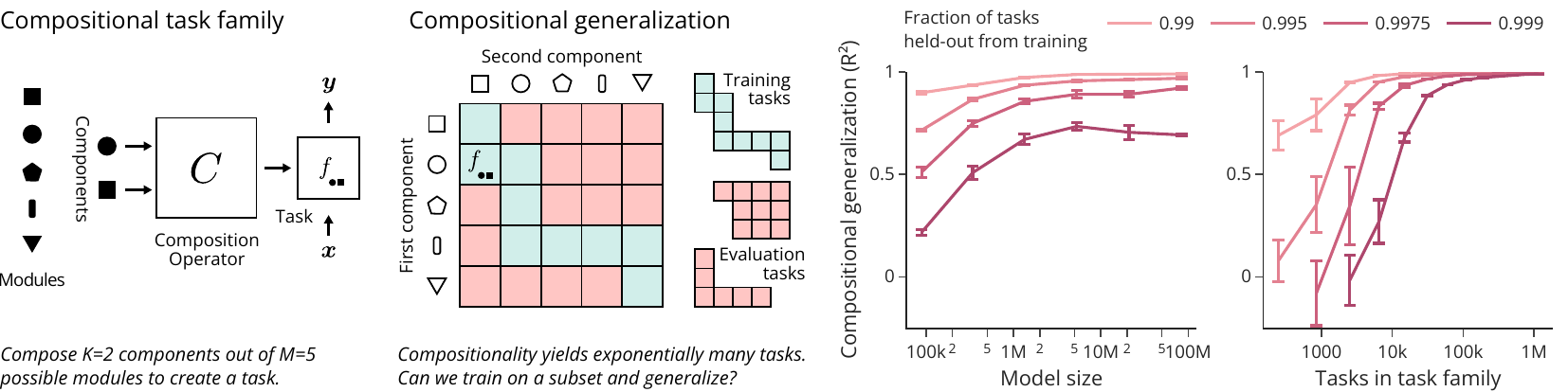}
    \vspace{-1em}
    \caption{\textbf{Scaling can lead to compositional generalization.} We consider compositional task families that compose $K$ out of $M$ modules into tasks, each of which is modeled as a function. This gives rise to an exponential number of $\mathcal{O}(M^K)$ tasks. We train standard feedforward networks on a subset of tasks and evaluate compositional generalization on held-out tasks. We find that scaling the size of the model and the data leads to compositional generalization.}
    \label{fig:graphical-abstract}
    \vspace{-1em}
\end{figure}
Here, we attempt to answer this question:
\begin{mdframed}[skipabove=1em, innertopmargin=1.75ex, innerbottommargin=1.5ex]
\text{Do neural networks compositionally generalize at scale?}
\end{mdframed}
\newpage
Our \textbf{main contributions} are as follows
\begin{itemize}[leftmargin=*]
  \item We demonstrate that standard multilayer perceptrons compositionally generalize on a variety of tasks as data and model size are scaled across task encodings if the training distribution sufficiently covers the task space.
  \item We prove that multilayer perceptrons can approximate a general class of compositional task families to arbitrary precision using only a linear number of neurons with respect to the number of task modules.
  \item We show that task constituents can be (linearly) decoded from the hidden activations of models that compositionally generalize, and demonstrate that this metric correlates with failures of image generation models to compose known concepts.
\end{itemize}

\section{Compositionality and compositional generalization}

We begin by formalizing compositionality and compositional generalization with the goal of capturing a variety of compositional data types including visual scenes, abstract reasoning and behavior policies.

\subsection{Compositional task family}

Specifically, we will consider \textit{compositional task families} that specify a generative procedure over tasks with shared compositional structure.
In a similar vein to \citep{elmoznino_complexity-based_2025}, our definition uses algorithmic complexity theory, in particular the notion of Kolmogorov complexity, see \citep{li_introduction_2019} for a formal treatment.
This definition is a modified version of the definition introduced by \citep{schug_meta-learning_2025}.\footnote{For the sake of simplicity, we are stating the definition slightly informally. For the asymptotic behavior over $K$ used in condition (ii) to be defined, we are technically considering a family of compositional task families.}

\begin{definition}[Compositional task family]
\label{def:compositional-family}
A compositional task family is a tuple $\mathcal{T}=(C, p: \bm{z} \mapsto p(\bm{z}), p: \bm{x} \mapsto p(\bm{x}))$, where:
\begin{itemize}[leftmargin=*, parsep=1pt]
    \item The task constituent space is a set $\mathcal{Z} \subseteq \{\bm{z} \in [0,1]^M : 1 \leq \|\bm{z}\|_0 \leq K \leq M\}$ with corresponding task distribution $p(\bm{z})$. $K$ is the number of task components and $M$ is the number of task modules.
    \item A \textit{task} is a function $f_{\bm{z}}: \mathcal{X} \to \mathcal{Y}$ that labels data points $\bm{x} \sim p(\bm{x})$.
    \item The \textit{composition operator} is a mapping $C: \mathcal{Z} \to (X \to Y)$ that takes as input task constituents $\bm{z} \in \mathcal{Z}$ and maps them to a task, $C(\bm{z}) \coloneqq f_{\bm{z}}$ for which the following conditions hold:
\begin{enumerate}[leftmargin=*, label=(\roman*), parsep=1pt]
  \item $C(\bm{z}) \neq C(\bm{z}')$ for all $\bm{z} \neq \bm{z}'$ with $\bm{z}, \bm{z}' \in \mathcal{Z}$, i.e. $C$ is injective.
  \item The length of the shortest program that implements $C$ as a function of $K$ grows sub-exponentially in $K$.
\end{enumerate}
\end{itemize}
\end{definition}
In the discrete case, where $\mathcal{Z} \subseteq \{\bm{z} \in \{0,1\}^M : 1 \leq \|\bm{z}\|_0 \leq K \leq M\}$ is restricted to the set of binary, $K$-hot vectors, Definition~\ref{def:compositional-family} essentially states that a compositional task family compactly captures exponentially many tasks.
Condition (i) ensures that all task components functionally enter the composition, while condition (ii) excludes the case where compositions are purely context-sensitive, ensuring that there is shared structure between tasks.
For a more detailed discussion of this definition, please refer to  \citep{schug_meta-learning_2025}.

The notion of a task is used in a general sense here and allows to capture different types of compositional data.
For instance, a task could refer to a visual scene, where modules are the set of possible objects and the composition operator renders a selection of such objects into a scene.
Similarly, a task could refer to a behavior policy, where modules consist of different reward functions, a subset of which is combined by the composition operator to induce an optimal policy.

\subsection{Task encoding}

Before we can continue to define compositional generalization using Definition~\ref{def:compositional-family}, we must first specify how to present the model with information about its current task, as captured by the task constituents $\bm{z}$.
In practice, such a task description might not be the task constituents themselves, but rather some encoding thereof.
For example, a task could be described through a natural language instruction or by presenting example data points $(x_i, f_{\bm{z}}(x_i))_i$.
To model this aspect, we define the task encoder as the mapping
\begin{align*}
\varphi: (\mathcal{Z}, \mathbb{N}) \rightarrow \mathcal{Z}'
\end{align*}
that maps task constituents $\bm{z} \in \mathcal{Z}$ and a random seed $r \in \mathbb{N}$ to a task encoding.
Throughout the paper, we mostly focus on settings where the task is unambiguously specified, i.e. where the task encoding $\varphi$ is information-preserving and therefore injective.

\subsection{Compositional generalization}

With Definition~\ref{def:compositional-family} at hand, we can now formalize compositional generalization for a model that learns to perform tasks from a compositional task family, given a task encoding $\varphi$.
This definition is a slightly modified version of the definition presented in \citep{schug_meta-learning_2025}.

\begin{definition}[Compositional generalization]
\label{def:compositional-generalization}
A model parameterized by $\bm{\theta}$ is said to compositionally generalize with respect to the compositional task family $\mathcal{T} = (\mathcal{Z}, C, p(\bm{z}), p(\bm{x}))$ if there exists a discrete training distribution $\bm{z}\mapsto p^{\mathrm{train}}(\bm{z})$ with finite support such that the number of points in the support grows sub-exponentially in $K$ and it holds that
\begin{align*}
  \bm{\theta}^* &\in \argmin_{\bm{\theta}} \mathbb{E}_{\bm{z} \sim p^{\mathrm{train}}(\bm{z})} \left [ \mathbb{E}_{\bm{x} \sim p(\bm{x})} \left [ l(\bm{\theta}, \bm{x}, \bm{z}) \right ] \right ] \\
  \Rightarrow \bm{\theta}^* &\in \argmin_{\bm{\theta}} \quad \mathbb{E}_{\bm{z} \sim p(\bm{z})} \left [ \mathbb{E}_{\bm{x} \sim p(\bm{x})} \left [ l(\bm{\theta}, \bm{x}, \bm{z}) \right ] \right ],
\end{align*}
where $l(\bm{\theta}, \bm{x}, \bm{z}) = l \left (f_{\bm{z}}(\bm{x}), g_{\bm{\theta}}(\bm{x}, \varphi(\bm{z}) \right )$ is a loss function that measures the discrepancy between model predictions and the outputs of a task $f_{\bm{z}}(\bm{x})$ for a given datum $\bm{x}$ and task encoding $\varphi(\bm{z})$.
\end{definition}
Note, that we are not considering fixed-size datasets but are sampling directly from the data distribution which reflects the nowadays common single-epoch training regime of large-scale foundation models \citep[e.g.,][]{brown_language_2020}.

\subsection{Hyperteachers: A general class of compositional task families}
\label{sec:hyperteacher}
For the purpose of this study, it will be useful to instantiate a concrete but nevertheless general class of compositional task families according to Definition~\ref{def:compositional-family}.
Given that neural networks are flexible function approximators and thus able to cover a wide range of behaviors, we parameterize both the composition operator as well as the task functions using neural networks.
The resulting system, a composable neural network that generates another neural network, can be interpreted as a hypernetwork \citep{ha_hypernetworks_2017}.
Indeed, hypernetworks have previously been used to study compositional generalization \citep{schug_discovering_2024}.

Following \citep{schug_discovering_2024}, we consider a linear hypernetwork that sums $K$ out of $M$ possible weight matrices to parameterize a single hidden layer task network.
We define the task constituent space to be $\mathcal{Z} \subseteq \{\bm{z} \in \{0.5, 0.6, \dots ,1.0\}^M : 1 \leq \|\bm{z}\|_0 \leq K \leq M\}$ and the composition operator as
\begin{align}
\label{eq:hyperteacher}
C(\bm{z}) = \bm{x} \mapsto \bm{\Omega} \;\mathrm{ReLU}\left (\sum_{k: z_k \neq 0} z_k \bm{\Theta}_k \bm{x} \right),
\end{align}
where we have $M$ sets of neural network parameters $\{ \bm{\Theta}_{m} \in \mathbb{R}^{I \times H} \}_{m=1}^M$ with $I$ input neurons and $H$ hidden neurons, $\bm{\Omega} \in \mathbb{R}^{H \times O}$ is a shared readout projection with $O$ output neurons and we sample $\bm{x} \in \mathbb{R}^I$ from the uniform distribution $p(\bm{x}) = \mathcal{U}[-1, 1]^I$.
Each module $\bm{\Theta}_m$ also has an associated bias vector which we omit here for conciseness.
The resulting task functions, $f_{\bm{z}}: \mathbb{R}^I \to \mathbb{R}^O$, are thus single hidden layer ReLU networks.

\section{Scaling can lead to compositional generalization}
\label{sec:scaling}

\begin{figure}[t]
  \includegraphics[width=0.5\textwidth]{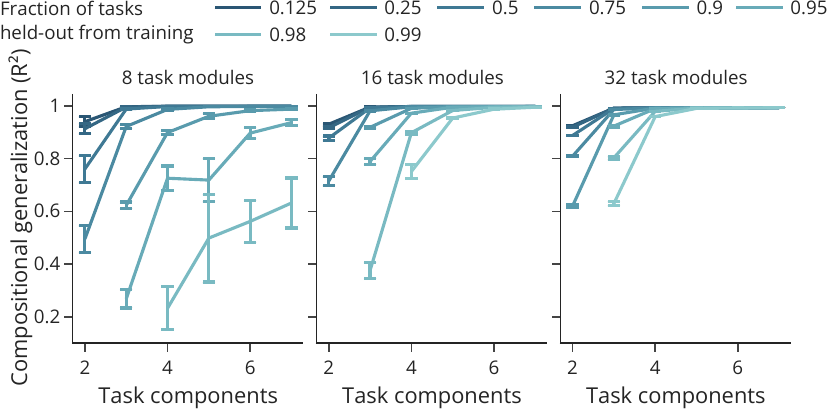}
  \includegraphics[width=0.5\textwidth]{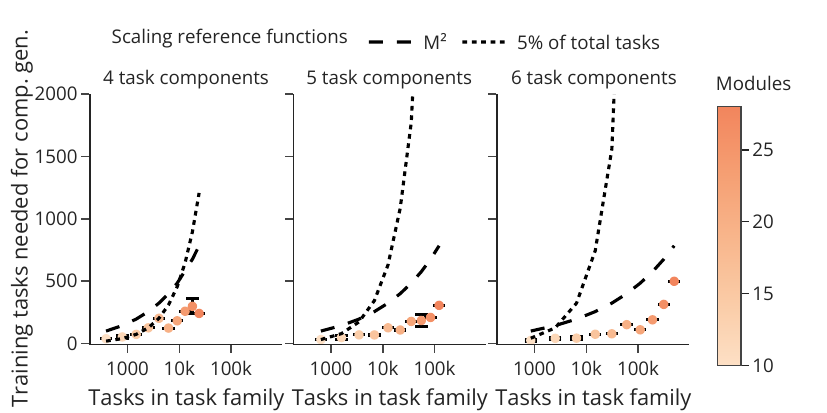}
  \includegraphics[width=\textwidth]{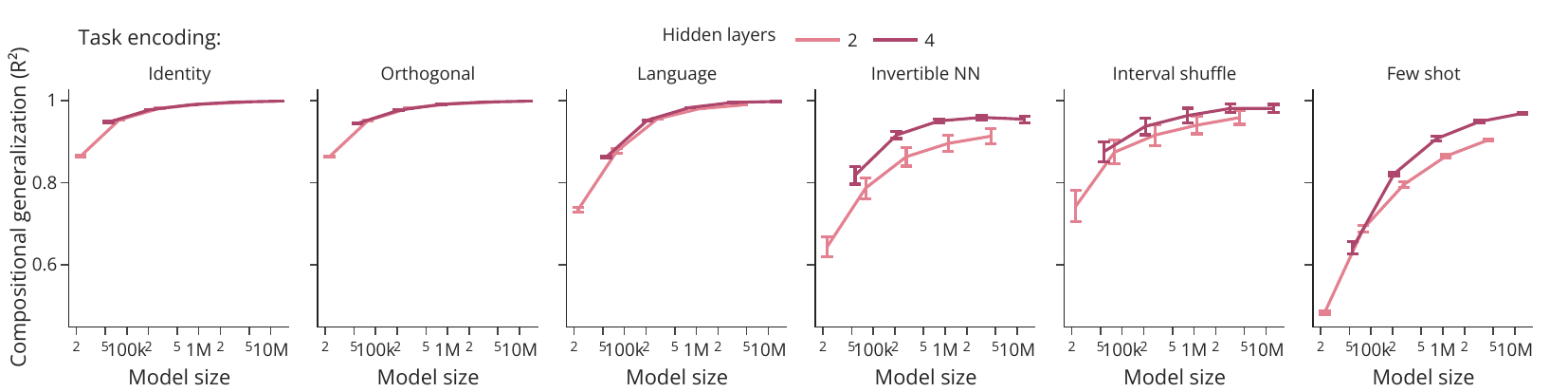}
  \vspace{-1em}
  \caption{\textbf{Scaling data and model size leads to compositional generalization.} \textit{\textbf{Top-left}} Scaling the number of training tasks by increasing the number of modules, task components or decreasing the fraction of tasks held-out from training leads to compositional generalization on the hyperteacher task family. \textit{\textbf{Top-right}} The number of training tasks required to achieve compositional generalization, here defined as a $R^2>0.95$, scales sub-exponentially as the number of total tasks in the task family grows. \textit{\textbf{Bottom}} Scaling model size by increasing the number of hidden neurons and the number of hidden layers leads to compositional generalization on the hyperteacher across different task encodings. Error bars denote SEM over three seeds.}
  \label{fig:scaling}
  \vspace{-2em}
\end{figure}

In the following, we will investigate if and under what circumstances standard multilayer perceptrons compositionally generalize.
We will show that simply scaling the number of tasks in the training distribution as well as scaling model size leads to compositional generalization.
The results presented in this section focus on various parameterizations of the hyperteacher compositional task family presented in Section~\ref{sec:hyperteacher}.
\textbf{We reproduce all findings of this section on the compositional preference task family introduced by \citep{schug_discovering_2024}.}
This task family requires learning optimal policies in a grid world with compositional reward functions.
We present the corresponding results in Appendix~\ref{appsec:preference-grid} in Figures~\ref{appfig:scaling-preference}, \ref{appfig:task-support-preference}, \ref{appfig:task-decoding-preference} and Table~\ref{apptab:task-encodings-preference}.
For additional experimental details, please refer to Appendix~\ref{appsec:experimental-details}.

While there exist various architectures specialized for compositionality, here we are interested in understanding if a standard fully connected neural network, a multilayer perceptron, can compositionally generalize.
Fully connected layers are common building blocks of virtually all standard architectures, including transformers or recurrent neural networks.
Specifically, we consider a multilayer perceptron with ReLU nonlinearities that accepts the concatenation of $\bm{x}$ and $\varphi(\bm{z}, r)$ as input.
\begin{center}
  \input{Figures/mlp.tikz}
\end{center}
In order to measure a model's ability to compositionally generalize, we will hold out tasks from training and evaluate the model's performance on these tasks.

\subsection{Scaling the number of compositional tasks leads to compositional generalization}

To investigate the main question of how scale affects compositional generalization, we will vary both data and model size.
The former can be accomplished along two different dimensions:
We can vary the total number of tasks in the compositional task family by changing both the number of modules $M$ and the number of components $K$, and we can vary the fraction of distinct tasks that are held-out from training.
Since the number of possible tasks grows exponentially, $\mathcal{O}(M^K)$, the compositional task families we consider can easily contain a very large number of distinct tasks.
The top-left of Figure~\ref{fig:scaling} shows that as we scale the number of tasks, compositional generalization improves.
Notably, the required number of training tasks to achieve compositional generalization grows more slowly than the total number of tasks as shown on the top-right of Figure~\ref{fig:scaling}.
This implies that compositional generalization with a sub-exponential number of tasks, as in Definition~\ref{def:compositional-generalization}, is indeed achievable at scale.
In Appendix~\ref{appsec:additional-results}, we further demonstrate that this scaling relationship is even more favorable for transformers and similarly holds for the compositional preference task family as well as a hyperteacher with a deep target network.

In addition to scaling the data, we would like to understand how scaling model size affects compositional generalization.
Shown at the bottom of Figure~\ref{fig:scaling}, we vary both the number of hidden layers and the size of the hidden layers for various possible task encodings $\varphi(\bm{z}, r)$ (more details on the task encodings will follow in Section~\ref{sec:task-encoding-results}).
We find that, given sufficient data, increasing model size consistently improves compositional generalization.
This is noteworthy, given that increasing model size in principle increases the capacity to memorize training tasks without capturing the underlying compositional structure required for compositional generalization.
As we will argue in the following however, these results can be interpreted as evidence that deep neural networks tend to prefer solutions of low algorithmic complexity \citep{wilson_deep_2025}.

\subsection{Complexity of generalizing solution dominates memorizing solution asymptotically}
\label{sec:generalizing-solution}

Memorizing all tasks of a compositional task family by definition requires exponential network capacity.
Intuitively, a solution that captures the underlying compositional structure and thus generalizes should be more efficient.
A priori, it is however not clear whether such a solution exists for a finite-sized, multilayer perceptron.
As we have argued before, hyperteachers can be regarded as a general class of compositional task families.
It is therefore instructive to consider whether a finite-sized multilayer perceptron can implement any hyperteacher without having to memorize the exponential number of possible tasks.
The following theorem answers this question in the affirmative.
\begin{theorem}
\label{thm:mlp-hyperteacher}
    Let $\left ( \bm{\Theta}_{m} \in \mathbb{R}^{I \times H} \right )$ be a sequence of uniformly bounded matrices.
    Then, for any $M\in\mathbb{N}$, $\varepsilon > 0$, and on any compact input set, $\mathcal{X} \times \mathcal{Z}$ with $\mathcal{Z} = \{ \bm{z}: \|\bm{z}\|_1 \leq 1 \}$, there exists a ReLU multilayer perceptron that approximates a hyperteacher to within $\varepsilon$ error in the $\|\cdot\|_{\infty}$ norm using $\mathcal{O}\left(\frac{1}{\sqrt\varepsilon}+M\right)$ neurons.
\end{theorem}
The corresponding constructive proof is presented in Appendix~\ref{appsec:proof-mlp-hyperteacher} along with an extension to hyperteachers with multiple layers.
Theorem~\ref{thm:mlp-hyperteacher} notably states that the number of neurons required for the generalizing solution scales linearly in the number of modules $M$.
Consistent with our experimental findings, this means that as $M$ grows, the simplicity of the fully generalizing solution will increasingly dominate the naive memorizing solution.

\subsection{Compositional generalization emerges across task encodings}
\label{sec:task-encoding-results}

\begin{figure}[t]
    \centering
    \includegraphics[width=\textwidth]{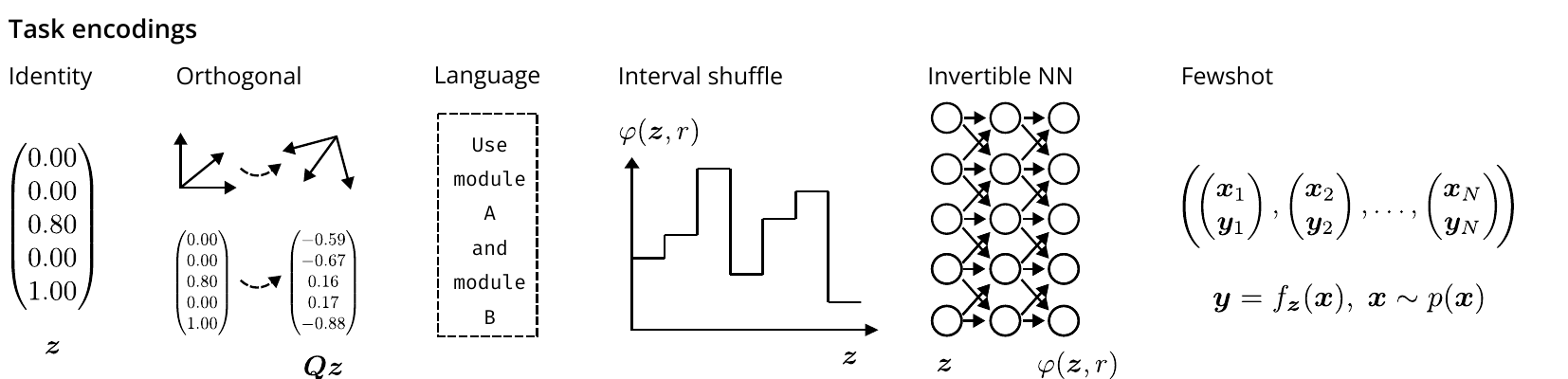}
    \vspace{-1em}
    \caption{\textbf{Task encodings.} Illustration of the different task encodings $\varphi(\bm{z}, r)$ used in Table~\ref{tab:task-encodings}. The first three encodings are linear with respect to the task constituents while the last three are nonlinear.}
    \label{fig:task-encoding}
\end{figure}

\begin{table}[t]
    \centering
    \caption{\textbf{Compositional generalization emerges across task encodings.} Comparison of the decodability of the task constituents from the hidden activations of the third layer (Task decoder) and compositional generalization performance for various task encodings for a hyperteacher with $M=16$, $K=3$. We additionally show the linear decodability of the task constituents directly from the task encoding itself (Input decoder), which allows to distinguish linear from nonlinear task encodings. $\pm$ SEM over three seeds.}
\label{tab:task-encodings}
\begin{tabular}{lrrr}
\toprule
Task encoding & Task decoder (R²) & Input decoder (R²) & Comp. gen. (R²) \\
\midrule
Identity & $0.95 \pm 0.012$ & $1.00 \pm 0.000$ & $1.00 \pm 0.000$ \\
Orthogonal & $0.96 \pm 0.002$ & $1.00 \pm 0.000$ & $1.00 \pm 0.000$ \\
Language & $0.99 \pm 0.000$ & $1.00 \pm 0.000$ & $1.00 \pm 0.000$ \\ \midrule
Invertible NN & $0.94 \pm 0.001$ & $0.56 \pm 0.000$ & $0.95 \pm 0.000$ \\
Interval shuffle & $0.96 \pm 0.011$ & $0.73 \pm 0.082$ & $0.98 \pm 0.010$ \\
Few shot & $0.90 \pm 0.004$ & $-0.23 \pm 0.008$ & $0.97 \pm 0.001$ \\
\bottomrule
    \end{tabular}
    \vspace{-1em}
\end{table}

We now turn to the question, to what extent the way in which the task is specified to the model matters for its ability to compositionally generalize.
Specifically, one might suspect that certain ways of encoding a task are better suited to leveraging compositional structure than others.
For instance, \citep{riveland_natural_2024} argue that language in particular encodes task structure in a way that is beneficial for learning compositional representations.
To study this question in the context of our setup, we experiment with different task encodings $\varphi(\bm{z}, r)$ illustrated in Figure~\ref{fig:task-encoding}.

Generally, we find that all task encodings lead to compositional generalization, including nonlinear encodings, although some require more model capacity as shown at the bottom of Figure~\ref{fig:scaling} and Table~\ref{tab:task-encodings} (also see Table~\ref{apptab:task-encodings-preference} in the appendix for corresponding results on the compositional preference task family).
Specifically, we observe no benefit of directly using the identity task encoding $\varphi(\bm{z}, r)=\bm{z}$, over 
a random but fixed orthogonal projection $\varphi(\bm{z}, r)=\bm{Q} \bm{z}$ where $\bm{Q} \in \mathbb{R}^{M \times M}$ is an orthogonal matrix.
In the same way, encoding each task through a language instruction poses no issues.
The latter is consistent with the observation that the task constituents can be linearly decoded from such instructions.
Interestingly, even nonlinear encodings such as specifying the task through examples (denoted fewshot), via an invertible neural network or from the highly nonlinear interval shuffle function (see Algorithm~\ref{alg:interval-shuffle} in the appendix for a definition) lead to compositional generalization if the model size is sufficiently large.
One possible explanation for these findings is that regardless of the task encoding, the model internally infers the task constituents up to a linear transformation after which Theorem~\ref{thm:mlp-hyperteacher} guarantees that a generalizing solution that scales linearly in the number of modules $M$ exists.

To verify this hypothesis, we train a linear decoder to predict the task constituents based on the hidden activations of the model solving the task.
We report the ability of this task decoder to predict the task constituents on the held-out compositional generalization tasks in Table~\ref{tab:task-encodings}.
Indeed, we find that also for nonlinear task encodings the task constituents can be linearly decoded from the hidden activations providing evidence that the models internally linearize the task constituents to achieve compositional generalization.
We will expand on this finding in Section~\ref{sec:task-decodability}.

\subsection{The support of the training distribution needs to sufficiently cover the task space}
\label{sec:task-support}
\begin{figure}[t]
    \centering
    \includegraphics[width=\textwidth]{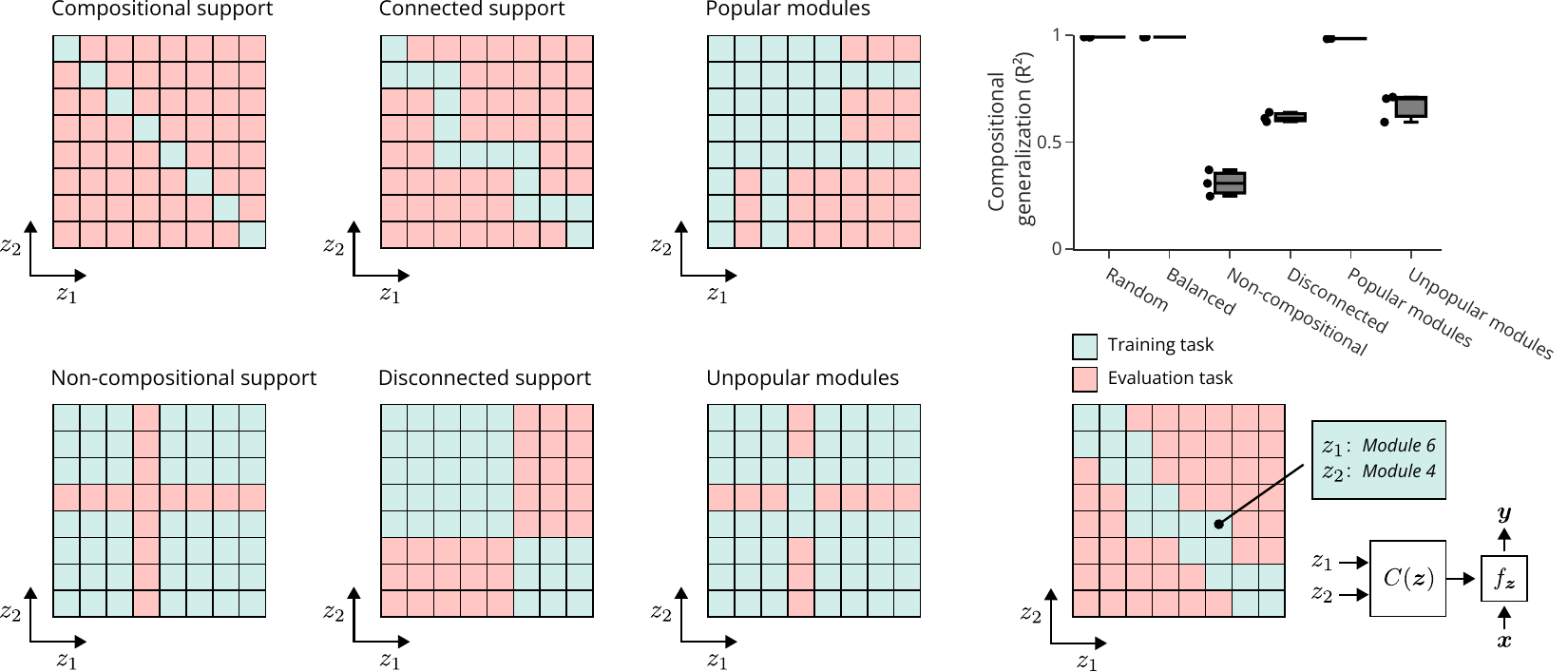}
    \vspace{-1em}
    \caption{\textbf{The support of the training distribution needs to sufficiently cover the task space.}: \textit{\textbf{Left}} Illustration of the different types of task support for the special case of a compositional task family with two components. Turquoise tiles denote module combinations that are part of the training distribution, red tiles are reserved for evaluation. \textit{\textbf{Right}} Compositional generalization as a function of the different types of task support on the hyperteacher for $M=16$ and $K=3$.}
    \label{fig:task-support}
    \vspace{-1em}
\end{figure}

In principle, it is easy to come up with degenerate training distributions that make compositional generalization impossible.
For instance, if a module is consistently absent from all training tasks, the model has no opportunity to learn this module and will generally fail to generalize to tasks that contain this module.
In this sense, the support of the training distribution $p^{\mathrm{train}}(\bm{z})$ needs to sufficiently cover the full constituent space for compositional generalization to succeed.
In this section we investigate how various conditions over the training support affect compositional generalization.

Prior work has studied how the training distribution affects compositional generalization \citep[e.g.,][]{wiedemer_compositional_2023,lippl_when_2024,schug_discovering_2024}.
Following \citep{wiedemer_compositional_2023} we refer to the condition of having non-zero support for each module in the training distribution as \textit{compositional support} .
For the class of hyperteacher task families, the additional condition of \textit{connected support} needs to be satisfied, which states that no subset of modules should appear solely in isolation from the rest.
\citep{schug_discovering_2024} show that in a teacher-student setting where both the teacher and the student are hypernetworks, this condition is required to guarantee compositional generalization.
We can extend this result to the more general case of any kind of student, including the ReLU multilayer perceptron.
The proof follows immediately by constructing examples of multiple different hyperteachers that have an identical training distribution if the training support is not connected.
Please refer to Appendix~\ref{appsec:proof-connectedness} for more details.

Figure~\ref{fig:task-support} illustrates the different types of training support we consider here as well as their effect on compositional generalization in the hyperteacher.
For a more detailed description for how each training support is constructed, please refer to Appendix~\ref{appsec:training-task-support}.
Our findings empirically confirm that violating compositional and connected support interferes with compositional generalization.
Interestingly, having a small set of popular modules who appear more frequently poses no issue for achieving compositional generalization.
The converse of having a small set of unpopular modules that are rarely encountered however does lead to a noticeable drop.
These findings are consistent with prior work that find that module imbalance hampers compositional generalization \citep{cong_attribute-centric_2023,okawa_compositional_2023}.
However, our experiments further suggest that such failures are due to underrepresentation of certain modules and not simply due to an asymmetry in module popularity.
Intuitively, if a module is seen only a constant number of times during training, it can be memorized with constant capacity.

\section{Task constituents are linearly decodable in models that compositionally generalize}
\label{sec:task-decodability}

Section~\ref{sec:task-encoding-results} has revealed that in cases where compositional generalization succeeds, the task constituents can be linearly decoded from the hidden activations.
This prompts the question whether models that succeed to compositionally generalize typically form an internal linear representation of the task constituents.
Generally speaking, we can show that for any model that compositionally generalizes to most tasks, the task constituents must be decodable.
\begin{theorem}[Decodability under compositional generalization]
\label{thm:decodability}
For any $\delta > 0$, assume we have a student $g$ that predicts labels $f_{\bm{z}}(\bm{x})=\bm{y}$ given a task encoding $\varphi(\bm{z},r)$ and inputs $\bm{x}$.
Then there exists a decoder map $\phi$ that decodes $\varphi(\bm{z},r)$ to $\bm{z}$ with probability at least $1-\sqrt{\delta}$, if:
\begin{enumerate}[leftmargin=*, label=(\roman*), parsep=1pt]
\item $\mathbb{P}_{\bm{z},\bm{x}}[g(\varphi(\bm{z},r), \bm{x})= C(\bm{z})(\bm{x})] > 1-\delta$
\item For each $\bm{z} \neq \bm{z}'$, $\mathbb{P}_{\bm{x}}[C(\bm{z})(\bm{x}) = C(\bm{z}')(\bm{x})] < 1-2\sqrt{\delta}$
    \end{enumerate}
\end{theorem}
We provide the proof in Appendix~\ref{appsec:proof-decodability}.
In practice, we observe an even stronger version of this statement, namely that the task constituents are \textit{linearly} decodable from the hidden activations in multilayer perceptrons that successfully compositionally generalize.

\subsection{Compositional generalization correlates with linear decodability of task constituents}
\label{sec:linear-decodability}
\begin{figure}[t]
    \centering
    \includegraphics[width=\textwidth]{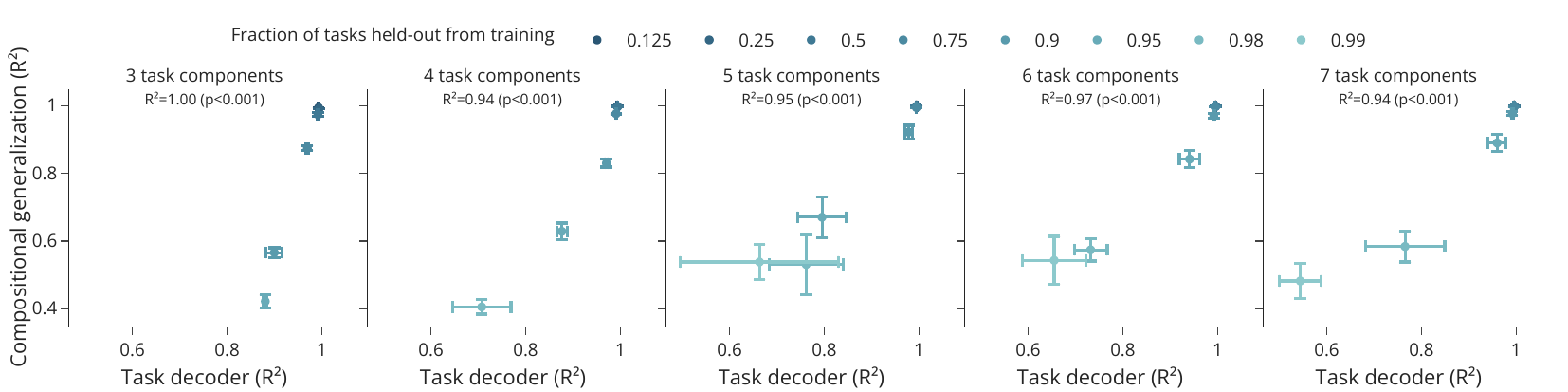}
    \includegraphics[width=\textwidth]{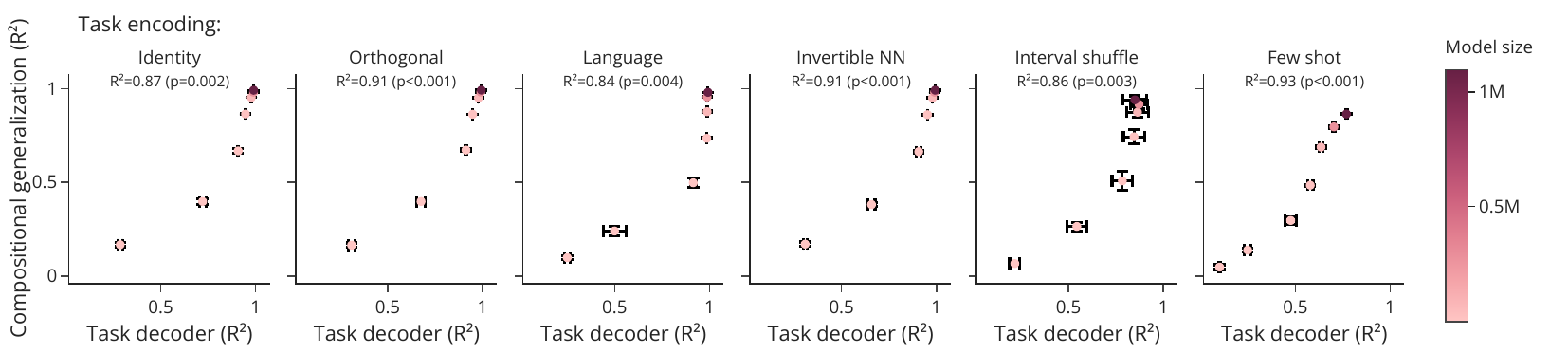}
    \vspace{-1em}
    \caption{\textbf{Compositional generalization correlates with linear decodability of task constituents.} \textit{\textbf{Top}} Relationship between linear decodability of the task constituents and compositional generalization across hyperteachers with $M=8$ modules and varying $K$. \textit{\textbf{Bottom}} Relationship between linear decodability of the task constituents and compositional generalization across different task encodings for varying model sizes on the hyperteacher with $M=16$, $K=3$. Error bars denote SEM over three seeds. \textit{\textbf{Top/Bottom}} We report the $R^2$ and corresponding p-value for an ordinary least square estimator in the facet titles.}
    \label{fig:task-decoding}
    \vspace{-1em}
\end{figure}

To further illuminate the connection between the observed linear decodability of the task constituents and compositional generalization, we attempt to decode the task constituents in models that (partially) fail to fully compositionally generalize.
Figure~\ref{fig:task-decoding} shows a remarkably clear correlation between decodability and compositional generalization across different data scales (top) and model sizes and task encodings (bottom).
Particularly interesting is the case where the unmodified task constituents are provided to the model, i.e. where the task encoding is the identity.
In this case, the task constituents are of course trivially linearly decodable from the input.
However, training the decoder on the hidden activations of deeper layers, this information is lost in networks that do not compositionally generalize.
In line with previous research, this implies that having access to a disentangled task representation is by itself not sufficient to achieve compositional generalization \citep{montero_lost_2022, kobayashi_when_2024}.

\subsection{Task constituents can be linearly decoded when image composition succeeds}

\begin{figure}[t]
    \centering
    \includegraphics[width=\textwidth]{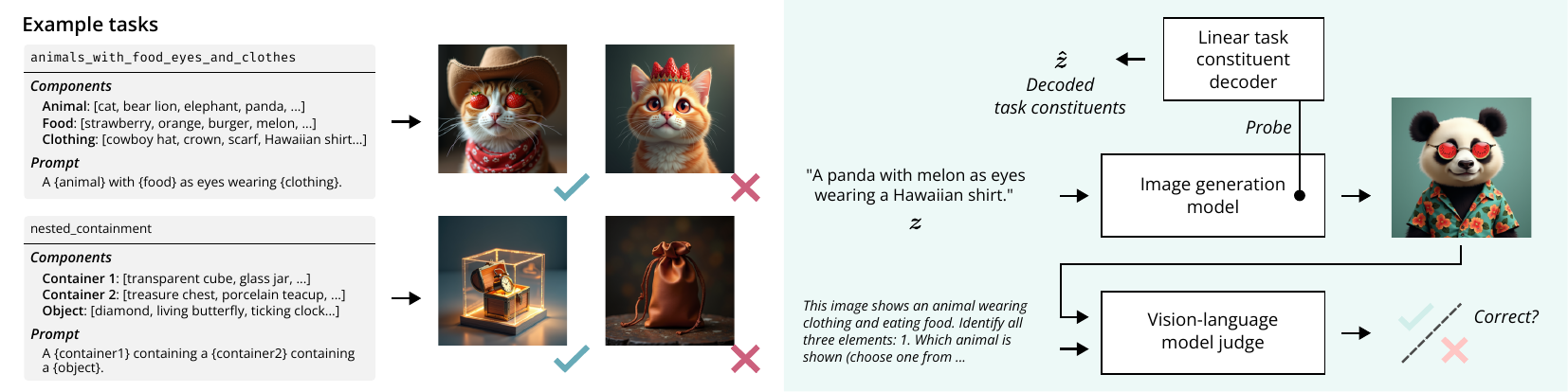}
    \includegraphics[width=\textwidth]{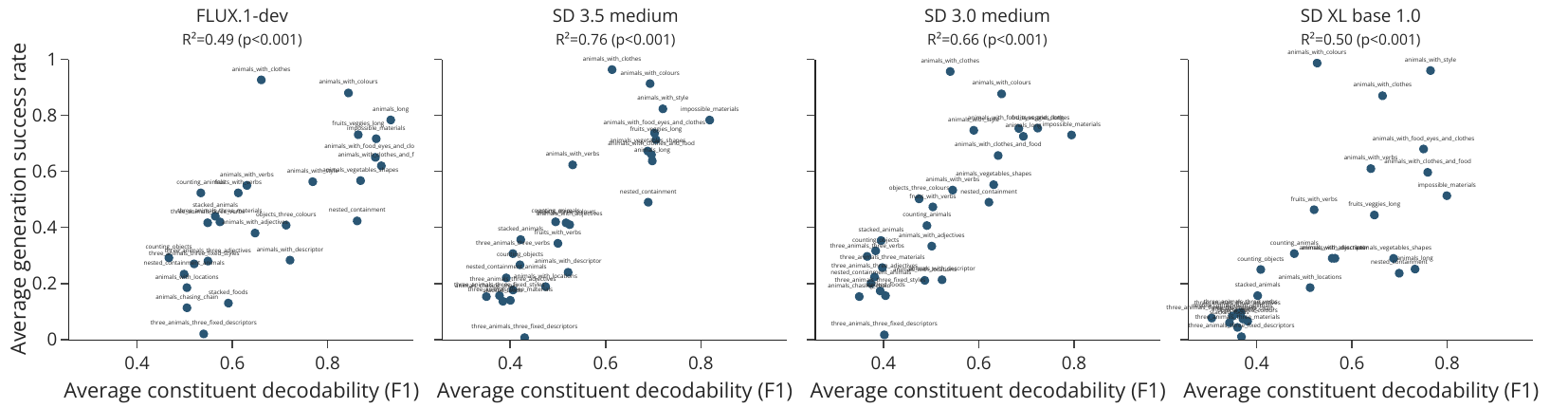}
    \vspace{-1em}
    \caption{\textbf{Task constituents can be linearly decoded when image composition succeeds.} \textit{\textbf{Top}} Example tasks to test compositionality of image generation models as well as an overview of the pipeline used to decode task constituents and judge model outputs. \textit{\textbf{Bottom}} Relationship between average constituent decodability and image generation success rate across image composition task families for four different image generation models. We report the $R^2$ and corresponding p-value for an ordinary least square estimator in the facet titles.}
    \label{fig:image-generation}
    \vspace{-1em}
\end{figure}

Finally, can we leverage the decodability of task constituents to gain insights into the successes and failures of image generation models at composing scenes from text prompts?
\footnote{Code for image composition experiments available at \url{https://github.com/florian-toll/compgen-vision}}
Image generation models have come a long way, displaying impressive abilities in creating novel compositions of known concepts \citep{okawa_compositional_2023}. 
Nevertheless, there are still systematic failure cases \citep{lewis_does_2024,wu_conceptmix_2024}.
Can such failures be related to the ability to infer the task constituents from the model's hidden activations?

To answer this question, we construct a large number of compositional tasks that require composing several concepts.
Two examples are shown in the top-left of Figure~\ref{fig:image-generation}.
For an extensive list of all the tasks we consider, as well as more details on the task construction, see Appendix~\ref{appsec:imgen-tasks}.
Using these tasks, we evaluate the ability of several diffusion-based image generation models to systematically generate image compositions \citep{black_forest_labs_black-forest-labsflux1-dev_2025, esser_scaling_2024}.
We use a vision-language model to label whether a given generation was successful and train a linear decoder to decode the task constituents based on the model's hidden activations.
The full pipeline is shown in the top-right of Figure~\ref{fig:image-generation}.
Please refer to Appendix~\ref{appsec:imgen} for further details.
The results of this analysis are shown at the bottom of Figure~\ref{fig:image-generation}.
We observe a clear correlation between the average task constituent decodability and the average generation success rate across models, with the relative task difficulty being similar across models, as shown in Figure~\ref{appfig:imgen-task-difficulty} of the Appendix.
This provides evidence that models which succeed at systematically composing known concepts into scenes form an internal representation of the task constituents.

\section{Related work}
\label{sec:related-work}

The study of compositional generalization in neural networks has a long and rich history, with early critiques highlighting the challenges of connectionist models to exhibit systematicity and compositionality \citep[e.g.,][]{fodor_connectionism_1988,smolensky_connectionism_1991, hadley_systematicity_1994,phillips_connectionism_1995} and numerous work that in response explored mechanisms for representing and processing structured information using distributed representations \citep[e.g.,][]{smolensky_tensor_1990,andreas_neural_2016}.
In recent years, theoretical progress has been made in showing that  compositional generalization can provably be achieved with neural networks in specific settings \citep{jarvis_specialization_2023,schug_discovering_2024, wiedemer_provable_2024,lee_why_2024,lippl_when_2024,brady_interaction_2025}.
This typically requires constraining the model architecture and the data generating process.
Consistent with our results, the statistics of the training data play a crucial rule in enabling compositional generalization \citep{wiedemer_compositional_2023,lippl_when_2024, schug_discovering_2024, brady_interaction_2025}.

We aim to complement this work by showing how scaling generic neural networks can lead to compositional generalization in the absence of stronger architectural constraints.
This is motivated by the finding that scaling neural networks can break the curse of dimensionality \citep{cagnetta_how_2024}, and consistently results in improvements in model performance \citep{kaplan_scaling_2020, hoffmann_training_2022} with new capabilities emerging as models are scaled up \citep{brown_language_2020, wei_emergent_2022, srivastava_beyond_2023}.
Compositional abilities in particular have therefore seemingly moved within grasp in practice \citep{zhou_least--most_2023, orhan_compositional_2022,furrer_compositional_2021, csordas_devil_2021, qiu_evaluating_2022,petty_impact_2024}.
However, even at larger scales, models often display a compositional generalization gap which does not close as scale increases \citep{dziri_faith_2023, press_measuring_2023, hosseini_not_2024,xu_large_2024, mirzadeh_gsm-symbolic_2025, wang_grokking_2024, yang_large_2024, berglund_reversal_2024, khandelwal_how_2025}, despite standard transformers showing compositionality in controlled settings \citep{lake_human-like_2023, schug_attention_2025, kumon_analyzing_2025}.

Image generation models in particular have made impressive leaps in their ability to create novel image compositions \citep{black_forest_labs_black-forest-labsflux1-dev_2025, betker_improving_2023, esser_scaling_2024, imagen-team-google_imagen_2024}.
Compositional abilities have been shown to emerge in such diffusion-based models on synthetic tasks in an order determined by the underlying data processes and with performance showing sudden emergence due to multiplicative dependencies \citep{okawa_compositional_2023,yang_swing-by_2025}.
Consistent with our finding that the task constituents are linearly decodable in models that successfully compositionally generalize, \citep{liang_how_2024} find that diffusion image generation models learn factorized representations on a number of synthetic tasks.
However, as the number of concepts that need to be composed grows, the performance of image generation models starts to deteriorate, showing the limits of their productivity to arbitrarily complex compositions \citep{lewis_does_2024,wu_conceptmix_2024}.

\section{Discussion}
\label{sec:discussion}

We have shown that simply scaling standard multilayer perceptrons can lead to compositional generalization challenging the position that stronger architectural priors are necessary to endow neural networks with compositionality \citep{boopathy_breaking_2025,battaglia_relational_2018, russin_systematicity_2020, du_compositional_2024}.
That being said, architectural priors do matter when it comes to data efficiency, as highlighted by the improved scaling of transformers over multilayer perceptrons (see Appendix~\ref{appsec:additional-results}).
Our findings also emphasize that the particular structure of the training data plays a critical role, demonstrating that not any type of scaling will lead to compositional generalization.

Interestingly, we find that when the training distribution is appropriately chosen, a wide range of task encodings support compositional generalization, including language but crucially also nonlinear task encodings such as specifying each task through examples.
Prior work posits that language compositionally structures neural representations and thereby aids rapid adaptation to novel compositional tasks \citep{riveland_natural_2024}.
Our results indicate that in fact any information-preserving mapping of the underlying task constituents suffices to achieve compositional generalization.
This might help explain why animals that do not use sophisticated languages can compositionally generalize \citep[e.g.,][]{tian_neural_2025}.

Surprisingly, regardless of whether task encodings are linear or nonlinear, we consistently find that the degree to which a model compositionally generalizes is related to the task constituents being linearly decodable from its hidden activations.
In particular, we show that this metric correlates with the success rate of image generation models at composing novel scenes from known constituents.
This finding suggests that, even though nonlinear representations would in principle be possible, successful compositional generalization in neural networks depends on linear representations of compositional structure.

\paragraph{Limitations}
While Theorem\ref{thm:mlp-hyperteacher} reveals the existence of a solution that enables compositional generalization without requiring an exponential number of neurons, identifying the theoretical conditions under which one can show that this solution is guaranteed to be discovered by stochastic gradient descent is an open question.
Prior results indicate that deep neural networks trained with stochastic gradient descent often display a preference towards simple solutions \citep{savarese_how_2019, chizat_implicit_2020, boursier_gradient_2022, wilson_deep_2025}.
In the context of compositional generalization, the complexity of the memorizing solution is by definition exponential in the number of modules which might help explain why empirically we observe the discovery of the generalizing solution which in contrast scales linearly in the number of modules.

We have focused on settings, where the task is fully specified through the task encoding.
In practice, a task specification might be ambiguous or incomplete, requiring models to handle uncertainty and infer the task at hand.
The strong in-context learning abilities of transformers that might allow extracting localized, and even cross-modal task vectors could play an important role in this context \citep{hendel_-context_2023,luo_vision-language_2025,yang_task_2025}.

More generally, scaling the data in a way that precisely controls the coverage of the training distribution requires an understanding of the true generative process underlying the compositional data.
For this reason, our scaling experiments rely on synthetic data generation that gives us the required experimental control.
On real-world data, the underlying generative process is mostly unknown and accordingly identifying training distributions that can support compositional generalization remains an open question.

\paragraph{Broader impacts}
This paper conducts foundational research aiming to illuminate under what circumstances neural networks compositionally generalize.
While we foresee no immediate negative societal impact, we hope that it may improve our understanding of this widely deployed technology.

\paragraph{Acknowledgments} We would like to thank Seijin Kobayashi, Angelika Steger, Jack Brady and Brenden Lake for vital discussions and fruitful feedback.
Simon Schug was supported by a Postdoc.Mobility grant (\texttt{P500PT\_225369}) from the Swiss National Science Foundation.

%% file: Figures/mlp.tikz
\begin{tikzpicture}[
    block/.style={draw, rectangle, minimum width=2cm, minimum height=1cm, thick},
    arrow/.style={-Triangle, thick, shorten >=3pt, shorten <=3pt}, %
    node distance=1cm
]

\node[block] (mlp) {ReLU MLP};
\node[left=of mlp, xshift=0.3cm, yshift=0.2cm] (x) {$\bm{x}$};
\node[left=of mlp, xshift=0.3cm, yshift=-0.2cm] (z) {$\varphi(\bm{z}, r)$};
\node[right=of mlp, xshift=-0.3cm] (y) {$\bm{y}$};

\draw[arrow] (x) -- (mlp.west|-x);
\draw[arrow] (z) -- (mlp.west|-z);
\draw[arrow] (mlp) -- (y);

\end{tikzpicture}

%% file: appendix.tex
\appendix

\section{Proofs}

\subsection{Connected support}
\label{appsec:proof-connectedness}

We briefly state the definition of connected support as introduced by \citep{schug_discovering_2024} using our notation.
\begin{definition}[Connected support]
\label{def:connected-support}
Let $\mathcal{T}=(C, p: \bm{z} \mapsto p(\bm{z}), p: \bm{x} \mapsto p(\bm{x}))$ be a compositional task family, and let $\bm{z}\mapsto p^{\text{train}}(\bm{z})$ be a distribution over $\mathcal{Z}$ with discrete support $\text{supp}(p^{\text{train}}) \subseteq \mathcal{Z}$.
We say that $\text{supp}(p^{\text{train}})$ is connected if the graph $G = (V, E)$ is connected, where $V = \{\bm{z} \in \mathcal{Z} \mid p^{\text{train}}(\bm{z}) > 0\}$ is the set of vertices and $E = \{(\bm{z}, \bm{z}') \in V \times V \mid \exists i \in \{1,2,\ldots,M\} \text{ such that } z_i = z'_i\}$ is the set of edges.
\end{definition}
In other words, two training task constituents are connected by an edge if and only if they share at least one element at the same position.
Section A.3 of \citep{schug_discovering_2024} provides an example that demonstrates how a student hypernetwork can perfectly fit the training distribution of a different teacher hypernetwork if the task support is compositional but disconnected.
This implies that there exist two distinct hyperteachers that have the same training distribution.
In such cases, any student, including the multilayer perceptron considered here, is not guaranteed to generalize even when perfectly fitting the training tasks.
To avoid those cases, connected support is generally required for the hyperteacher.

\subsection{Proof of Theorem~\ref{thm:mlp-hyperteacher}}
\label{appsec:proof-mlp-hyperteacher}

In the following, we will show that a multilayer perceptron can approximate a hyperteacher using a linear number of neurons in the number of task modules.
We first state the result for the single layer hyperteacher we primarily consider in the main text, before extending the result to hyperteachers with multiple layers.

\subsubsection{Single layer hyperteacher}
Let us recall the definition of the hyperteacher in Equation~\ref{eq:hyperteacher} with $M$ modules, $I$ input neurons, $H$ hidden neurons and $O$ output neurons,
\begin{align*}
(\bm{x}, \bm{z}) \mapsto \bm{\Omega} \;\mathrm{ReLU}\left(\sum_{m=1}^M \sum_{i=1}^I \bm{\Theta}_{m,i} z_m x_i\right),
\end{align*}
where $\{ \bm{\Theta}_{m} \in \mathbb{R}^{I \times H} \}_{m=1}^M$ are the modules, $\bm{\Omega} \in \mathbb{R}^{H \times O}$ is a readout projection and $\bm{z} \in [0,1]^M$ are the task constituents.
We are restating Theorem~\ref{thm:mlp-hyperteacher} here for convenience.

\begin{theorem*}[\ref{thm:mlp-hyperteacher}]
    Let $\left ( \bm{\Theta}_{m} \in \mathbb{R}^{I \times H} \right )$ be a sequence of uniformly bounded matrices.
    Then, for any $M\in\mathbb{N}$, $\varepsilon > 0$, and on any compact input set, $\mathcal{X} \times \mathcal{Z}$ with $\mathcal{Z} = \{ \bm{z}: \|\bm{z}\|_1 \leq 1 \}$, there exists a ReLU multilayer perceptron that approximates a hyperteacher to within $\varepsilon$ error in the $\|\cdot\|_{\infty}$ norm using $\mathcal{O}\left(\frac{1}{\sqrt\varepsilon}+M\right)$ neurons.
\end{theorem*}

\begin{proof}
    We will prove the statement, by providing an explicit construction of a ReLU multilayer perceptron that approximates a hyperteacher.
    The construction relies on two key lemmas: in Lemma~\ref{lem:square}, we show that the square function can be approximated in a ReLU multilayer perceptron, which we use in Lemma~\ref{lem:multiply} to show how the multiplication of two numbers can be approximated.
    Based on these building blocks, we then provide the construction of the full hyperteacher approximation.
    \begin{lemma}
    \label{lem:square}
        For any $\varepsilon > 0$, there exists a ReLU multilayer perceptron that approximates  $$ \mathrm{square} := \left\{ \begin{array}{clc}
            [0,1] & \to &\mathbb{R}\\
            x & \mapsto &x^2
        \end{array}\right.$$
         to within $\varepsilon$ error in the $\|\cdot\|_{\infty}$ norm using $\mathcal{O}(1/\varepsilon^2)$ neurons.
    \end{lemma}
    \begin{proof}
        Let us first consider how any function $f$ can be approximated by a piecewise linear function $L$ matching $f$ on $x_i \coloneqq \frac{i}{n}$ for $i \in \{0, 1, \dots, n\}$, where $n$ is a fixed integer. We define \begin{align*}
            L:x \mapsto f(0) + \sum_i \Big(&\frac{f(x_{i+1})-f(x_i)}{x_{i+1}-x_i}\mathrm{ReLU}(x-x_i) - \frac{f(x_{i})-f(x_{i-1})}{x_i-x_{i-1}}\mathrm{ReLU}(x-x_{i-1}) \Big)
        \end{align*}
        where $\frac{f(x_{0})-f(x_{-1})}{x_0-x_{-1}}$ is set to 0 by convention.
        
        It can easily be verified that $L$ is linear on any interval $[x_i, x_{i+1}]$ for $i \in \{0, 1, \dots, n\}$, and coincides with $f$ on $\left \{\frac{i}{n}, i \in \{0, 1, \dots, n\} \right \}$.
        We now want to bound $\|f-L\|_{\infty}$.
        While more general results can be shown for $f \in \mathcal{C}^2$, for $f = \mathrm{square}$ we can derive our result in the following simple way:   
        for any $x$ on any interval $[x_i, x_{i+1}]$, $$L(x) = \frac{f(x_{i+1})-f(x_i)}{x_{i+1}-x_i}(x-x_i) + f(x_i)$$
        so \begin{align*}
            |L(x) - f(x)| &= \left|\frac{x_{i+1}^2-x_i^2}{x_{i+1}-x_i}(x-x_i) + x_i^2 - x^2\right|\\
            &= (x-x_i)(x_{i+1}-x)\\
            &\leq \varepsilon &&\text{for } n\geq 1/\sqrt{\varepsilon} 
        \end{align*}
    \end{proof}
    An immediate corollary is that this result holds on any fixed bounded set. 
    We now show how to multiply two numbers using a ReLU multilayer perceptron:
    \begin{lemma}
        \label{lem:multiply}
        For any $\varepsilon > 0$, there exists a ReLU multilayer perceptron that approximates  $$\mathrm{multiply} :=\left\{ \begin{array}{clc}
            [0,1]^2 & \to &\mathbb{R}\\
            (x,y) & \mapsto &xy
        \end{array}\right.$$ to within $\varepsilon$ error in the $\|\cdot\|_{\infty}$ norm using $\mathcal{O}(1/\varepsilon^2)$ neurons.
    \end{lemma}
    \begin{proof}
        Using the polarization identity $xy = \frac{(x+y)^2-(x-y)^2}{4}$, we can approximate $\mathrm{multiply}(x,y)$ with a ReLU multilayer perceptron as follows:
        $$(x,y) \mapsto (x+y, x-y) \mapsto ((x+y)^2, (x-y)^2) \mapsto xy$$
    \end{proof}
    Again, this result holds on any fixed bounded set.

    We can now state a construction for a multilayer perceptron that approximates the preactivation of the hyperteacher, namely $(\bm{x}, \bm{z}) \mapsto \sum_{m=1}^M \sum_{i=1}^I \bm{\Theta}_{m,i} z_m x_i$ using a linear number of neurons in the number of task modules.

    \begin{lemma}
        \label{lem:layer}
        Let $\left ( \bm{\Theta}_{m} \in \mathbb{R}^{I \times H} \right )$ be a sequence of uniformly bounded matrices.
        Then, for any $M\in\mathbb{N}$, $\varepsilon > 0$, and on any compact input set, $\mathcal{X} \times \mathcal{Z}$ with $\mathcal{Z} = \{ \bm{z}: \|\bm{z}\|_1 \leq 1 \}$, there exists a ReLU multilayer perceptron that approximates $(\bm{x}, \bm{z}) \mapsto \sum_{m=1}^M \sum_{i=1}^I \bm{\Theta}_{m,i} z_m x_i$ to within $\varepsilon$ error in the $\|\cdot\|_{\infty}$ norm using $\mathcal{O}\left(\frac{1}{\sqrt\varepsilon}+M\right)$ neurons.
    \end{lemma}
    \begin{proof}
        We construct our multilayer perceptron layer by layer, tracking the number of neurons required for each layer and the error it incurs in the approximation:
        \begin{itemize}[leftmargin=*]
            \item The first layer copies the input $\bm{x}$ and computes $(\bm{x}, \bm{z}) \mapsto (\sum_{m=1}^{M} z_m \Theta_{m,i,h})_{i, h}$ for each $i \in \{1, \ldots, I\}$ and $h \in \{1, \ldots, O \}$, canceling the ReLUs by adjusting the biases accordingly given that the input is bounded. This requires $\mathcal{O}(M)$ neurons.
            \item Using Lemma~\ref{lem:multiply} with error $\frac{\varepsilon}{I}$, the next three layers multiply $\sum_{m=1}^{M} z_m \Theta_{m,i,h}$ by $x_i$ for each $i \in \{1, \ldots, I\}$ and $h \in \{1, \ldots, O \}$, using $\mathcal{O}(1/\sqrt\varepsilon)$ neurons.
            \item The next layer sums over $i$ for each $h \in \{1, \ldots, O \}$, again canceling  the ReLU nonlinearity by adjusting the biases.
            This requires $M$ neurons and incurs a final error of at most $\varepsilon$ since each output neuron sums the error coming from $I$ neurons.
        \end{itemize}
        In total, this construction uses $\mathcal{O}(M) + \mathcal{O}\left(\frac{1}{\sqrt\varepsilon}\right) + \mathcal{O}(M) = \mathcal{O}\left(\frac{1}{\sqrt\varepsilon}+M\right)$ neurons.
    \end{proof}

    Since $\mathrm{ReLU}$ is a contracting function, the output error after applying the $\mathrm{ReLU}$ activation is also $\varepsilon$.
    Finally, we can straightforwardly apply the output projection $\bm{\Omega}$ using a constant number of neurons, scaling the error by a constant.
\end{proof}

\subsubsection{Multilayer hyperteacher}
We now consider a multilayer hyperteacher, where a linear hypernetwork configures a multilayer perceptron with $L$ layers, $H_l$ neurons for layer $l$, $H_0=I$ input neurons and $H_L=O$ output neurons.
In this case, we have $C(\bm{z}) \coloneqq \bm{\Omega} \bm{h}_{L}$, a sequence of hidden layers,
\begin{align*}
    \bm{h}_{l+1} = \mathrm{ReLU} \Big(\underbrace{\sum_{m=1}^M z_m \bm{\Theta}_{l, m} }_{=: \bm{W}_l(\bm{z})}\bm{h}_l \Big),
\end{align*}
with input $\bm{h}_0 = \bm{x}$, a sequence of modules for each layer, $\left\{ \bm{\Theta}_{l,m} \in \mathbb{R}^{H_l \times H_{l+1}} \right\}_{\substack{l=1,\ldots,L \\ m=1,\ldots,M}}$, and output projection $\bm{\Omega} \in \mathbb{R}^{H_{L-1} \times O}$.

\begin{theorem}
    Let $\left ( \bm{\Theta}_{l, m} \in \mathbb{R}^{H_l \times H_{l+1}} \right )$ be a sequence of uniformly bounded matrices.
    Then, for any $M\in\mathbb{N}$, $\varepsilon > 0$, fixed $L$ and on any compact input set, $\mathcal{X} \times \mathcal{Z}$ with $\mathcal{Z} = \{ \bm{z}: \|\bm{z}\|_1 \leq 1 \}$, there exists a ReLU multilayer perceptron that approximates an $L$-layer hyperteacher to within $\varepsilon$ error in the $\|\cdot\|_{\infty}$ norm using $\mathcal{O}\left(\frac{1}{\sqrt\varepsilon}+M\right)$ neurons.
\end{theorem}
\begin{proof}
    First, observe that we can copy the task constituents $\bm{z}$ to each layer using $M$ neurons per layer.
    Because of this, we will assume that we have access to the task constituents at each layer.
    The proof then follows a similar approach to the proof of the single layer case of Theorem~\ref{thm:mlp-hyperteacher} but we must now consider how the error propagates through each layer.
    For the sake of simplicity, we will ignore the readout projection (i.e. assume $\bm\Omega=\bm{I}$), since this part of the proof is identical.
    
    We prove by induction on the number of layers $L$ that we can approximate $\bm{h}_L$ to within $\varepsilon$ error in the $\|\cdot\|_{\infty}$ norm with $\mathcal{O}\left(\frac{1}{\sqrt\varepsilon}+M\right)$ neurons.
    The base case for $L=0$ holds by definition, i.e. $\bm{h}_0 = \bm{x}$.
    Now, assume the result holds for an $L$-layer hyperteacher.
    Let $\varepsilon > 0$ and $\mathcal{X}$, $\mathcal{Z}$ be compact sets with $\mathcal{Z} = \{\bm{z}: \|\bm{z}\|_1 \leq 1 \}$.
    By the induction hypothesis, let $\hat{h}_L$ be a multilayer perceptron that approximates $\bm{h}_L$ on the compact set $\mathcal{X} \times \mathcal{Z}$, up to error $\varepsilon$ in the $\|\cdot\|_{\infty}$ norm.

    By Lemma~\ref{lem:layer}, let $g$ be a multilayer perceptron that approximates
    $(\bm{x}, \bm{z}) \mapsto \sum_{m=1}^M z_m \bm{\Theta}_{L+1,m} \bm{x}$
    on $\bar{B}_{\varepsilon} \times \mathcal{Z}$, where $\bar{B}_{\varepsilon}$ is the closed $\varepsilon$-ball around $\hat{h}_L(\mathcal{X} \times \mathcal{Z})$.
    Since the image of a compact set formed by a continuous map is compact and the closed epsilon ball around a compact set in a finite-dimensional space is compact, $\bar{B}_{\varepsilon}$ is also a compact set.
    
    Let us now show that $(\bm{x}, \bm{z}) \mapsto \mathrm{ReLU} \left ( g(\hat{h}_L(\bm{x}, \bm{z}), \bm{z}) \right )$ approximates $\bm{h}_{L+1}$ on $\mathcal{X} \times \mathcal{Z}$ up to error $\mathcal{O}(\varepsilon)$.
    We define $\bm{\Delta}_L \coloneqq \hat{h}_L(\bm{x}, \bm{z}) - \bm{h}_L$ for $L$ layers.
    Then, for $L+1$ layers it holds that,
    \begin{align*}
        \|\bm{\Delta}_{L+1}\|_{\infty} &= \|\mathrm{ReLU} (g(\bm{h}_L+\bm{\Delta}_L, \bm{z})) - \bm{h}_{L+1} \|_{\infty}\\
        &= \|g(\bm{h}_L+\bm{\Delta}_L, \bm{z}) - \bm{W}_L(\bm{z}) \bm{h}_L \|_{\infty}\quad\quad \text{(since } \mathrm{ReLU} \text{ is contracting)}\\
        &= \|g(\bm{h}_L+\bm{\Delta}_L, \bm{z})  - \bm{W}_L(\bm{z}) (\bm{h}_L+\bm{\Delta}_L) + \bm{W}_L(\bm{z}) \bm{\Delta}_L\|_{\infty}\\
        &\leq \|g(\bm{h}_L+\bm{\Delta}_L, \bm{z})  - \bm{W}_L(\bm{z}) (\bm{h}_L+\bm{\Delta}_L) \|_{\infty} + \| \bm{W}_L(\bm{z}) \|_{\infty \to \infty} \|\bm{\Delta}_L\|_{\infty}\\
        &= \varepsilon + \mathcal{O}(\varepsilon) = \mathcal{O}(\varepsilon),
    \end{align*}
    where $\| \cdot \|_{\infty \to \infty}$ is the operator norm induced by the $\|\cdot\|_{\infty}$ norm.
    The final error can be reduced to be at most $\varepsilon$ by adjusting the number of neurons by a constant factor.

    Finally, let us bound the number of neurons required for the full construction.
    We used $\mathcal{O}(M)$ neurons for copying $\bm{z}$ to each layer, $\mathcal{O}\left(\frac{1}{\sqrt{\varepsilon}} + M\right)$ neurons for the construction of $\hat{h}_L$, and $\mathcal{O}\left(\frac{1}{\sqrt{\varepsilon}} + M\right)$ neurons for the construction of $g$, which sums up to $\mathcal{O}\left(\frac{1}{\sqrt{\varepsilon}} + M\right)$ neurons.
\end{proof}

\subsection{Proof of Theorem~\ref{thm:decodability}}
\label{appsec:proof-decodability}
\begin{proof}
    Let $X, Z, R$ be independent random variables sampled according to their respective distributions.
    We define the event made of the set of pairs of task constituents $\bm{z}$ and seeds $r$ such that the accuracy is higher than $1-\sqrt{\delta}$:
    \begin{align*}
        A &\coloneqq \left[\mathbb{E}\left[\mathbbm{1}\{g(\varphi(Z, R), X)=C(Z)(X)\}|Z,R\right] \geq 1-\sqrt{\delta}\right]\\
        &=\left\{(\bm{z},r) \in \mathcal{Z} \times \mathbb{N} \mid \mathbb{P}[g(\varphi(\bm{z}, r), X)=C(\bm{z})(X)] \geq 1-\sqrt{\delta}\right\}
    \end{align*}

    where the expectation is taken over $X$.
    By Markov's inequality, and the first assumption, this set can't be too small. Indeed,
    \begin{align*}
        \mathbb{P}[\neg A] &< \frac{\mathbb{E}\left[\mathbbm{1}\{g(\varphi(Z, R), X) \neq C(Z)(X)\}\right]}{\sqrt\delta} && \text{ by Markov's inequality}\\
        &< \sqrt\delta && \text{ by the first assumption}
    \end{align*}

    We now show that for all $(\bm{z},r), (\bm{z'},r') \in A \subset\mathcal{Z} \times \mathbb{N}$
    $$\varphi(\bm{z},r)=\varphi(\bm{z'}, r') \implies \bm{z}=\bm{z'}.$$

    Indeed, let $(\bm{z},r), (\bm{z'},r') \in A \subset\mathcal{Z} \times \mathbb{N}$, with $\bm{\hat{z}}\coloneqq\varphi(\bm{z},r)=\varphi(\bm{z'}, r')$.

    We have, by definition of $A$, 
    \begin{align*}
        \mathbb{P}[g(\bm{\hat{z}}, X)=C(\bm{z})(X)] \geq 1-\sqrt{\delta}\\
        \mathbb{P}[g(\bm{\hat{z}}, X)=C(\bm{z'})(X)] \geq 1-\sqrt{\delta}
    \end{align*}

    so by union-bound, it holds

    $$\mathbb{P}[C(\bm{z})(X)=C(\bm{z'})(X)] \geq 1-2\sqrt{\delta}$$
    
    which is only possible if $\bm{z}=\bm{z'}$ given the second assumption.

    We now define $$\phi:\left\{\begin{array}{cccll}
        \mathcal{Z'} & \to & \mathcal{Z} & \\
         \bm{\hat{z}} & \mapsto & \bm{z} & \text{ for any } (\bm{z}, r) \in \varphi^{-1}(\bm{\hat{z}}) & \text{ if } \bm{\hat{z}} \in \varphi(A)\\
         \bm{\hat{z}} & \mapsto & \bm{0} && \text{ otherwise}
    \end{array}\right.$$

    $\phi$ is uniquely defined because of the previous property.
    
    We have $$\mathbb{P}[\phi(\varphi(Z,R))=Z] \geq \mathbb{P}[(Z,R)\in A] \geq 1 - \sqrt{\delta}$$
    which proves our statement.
\end{proof}

\newpage

\section{Additional results}
\label{appsec:additional-results}

\subsection{Compositional generalization on the preference grid}
\label{appsec:preference-grid}

The findings in Section~\ref{sec:scaling} and Section~\ref{sec:linear-decodability} focus on the hyperteacher task family.
In the following, we show that all findings can be reproduced on the compositional preference family introduced by \citep{schug_discovering_2024}.

\begin{figure}[t]
  \includegraphics[width=0.75\textwidth]{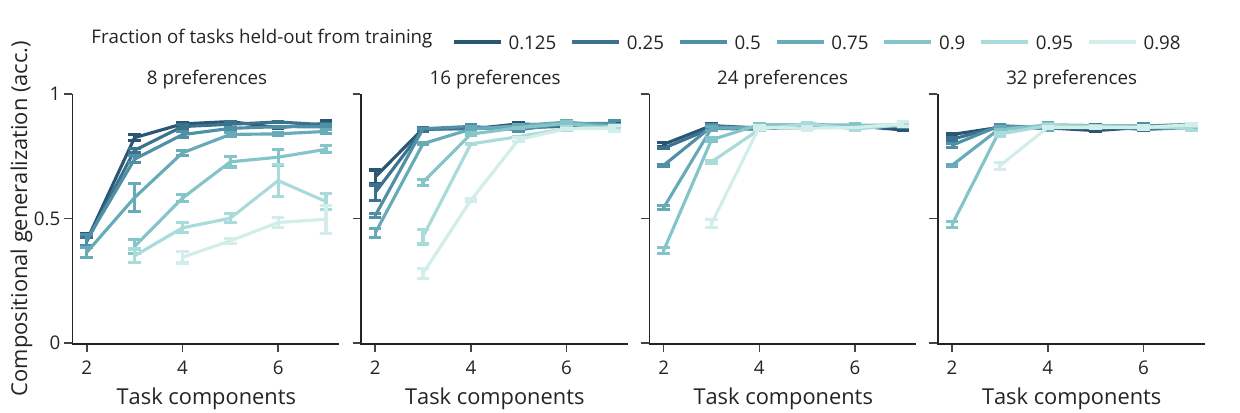}
  \includegraphics[width=\textwidth]{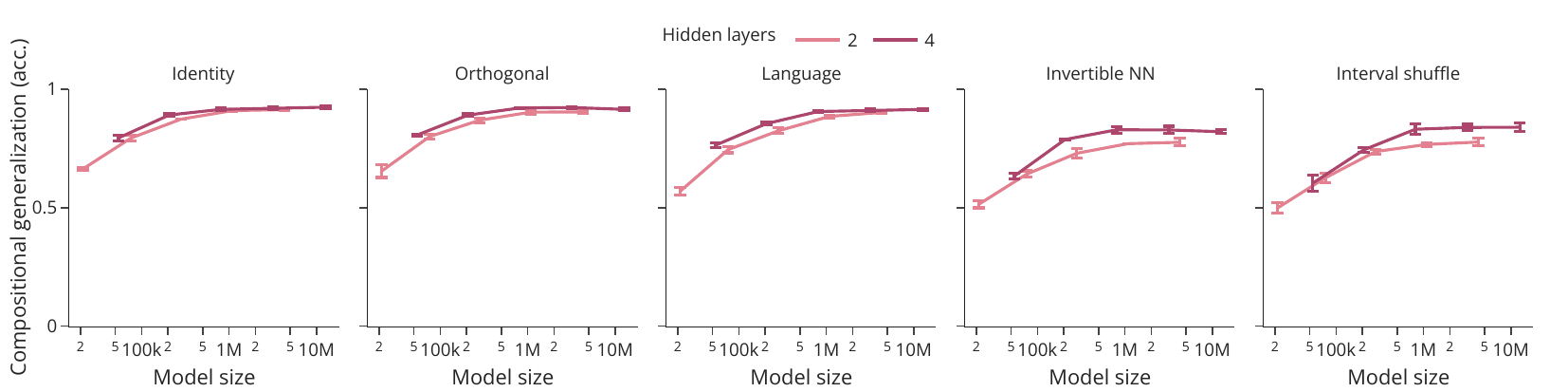}
  \caption{\textbf{Scaling data and model size leads to compositional generalization.} \textit{\textbf{Top}} Scaling the number of training tasks by increasing the number of modules, task components or decreasing the fraction of tasks held-out from training leads to compositional generalization on the compositional preference task family. \textit{\textbf{Bottom}} Scaling model size by increasing the number of hidden neurons or the number of hidden layers leads to compositional generalization on the compositional preference task family across different task encodings. Error bars denote SEM over three seeds.}
  \label{appfig:scaling-preference}
\end{figure}

\begin{figure}[t]
    \centering
    \includegraphics[width=0.5\textwidth]{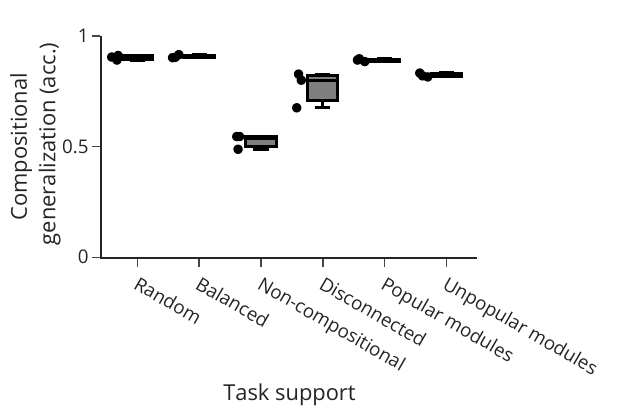}
    \caption{\textbf{The support of the training distribution needs to sufficiently cover the task space.}: Compositional generalization as a function of the different types of task support on the compositional preference task family for $M=16$ and $K=3$.}
    \label{appfig:task-support-preference}
\end{figure}

\begin{figure}[t]
    \centering
    \includegraphics[width=\textwidth]{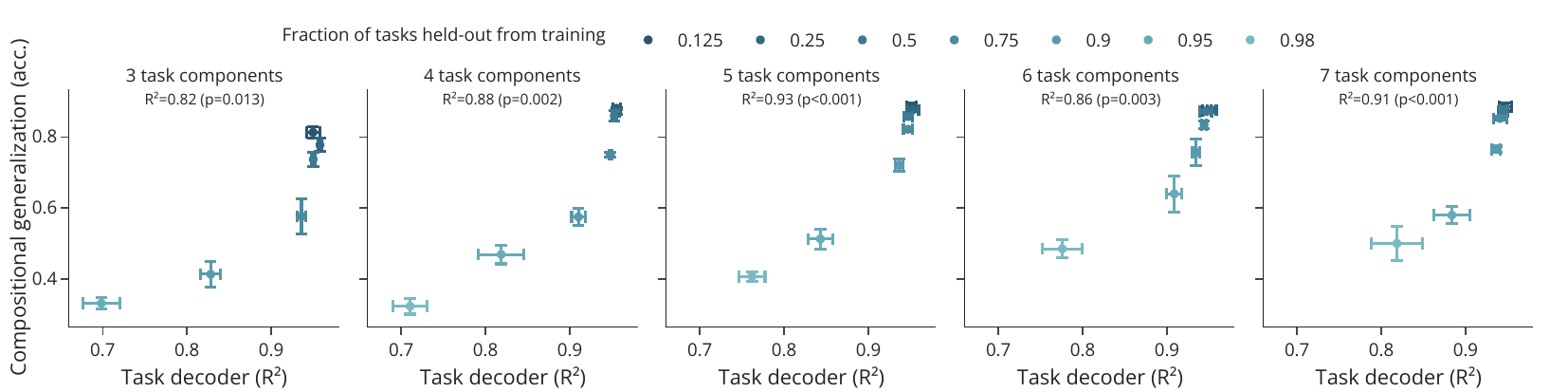}
    \includegraphics[width=\textwidth]{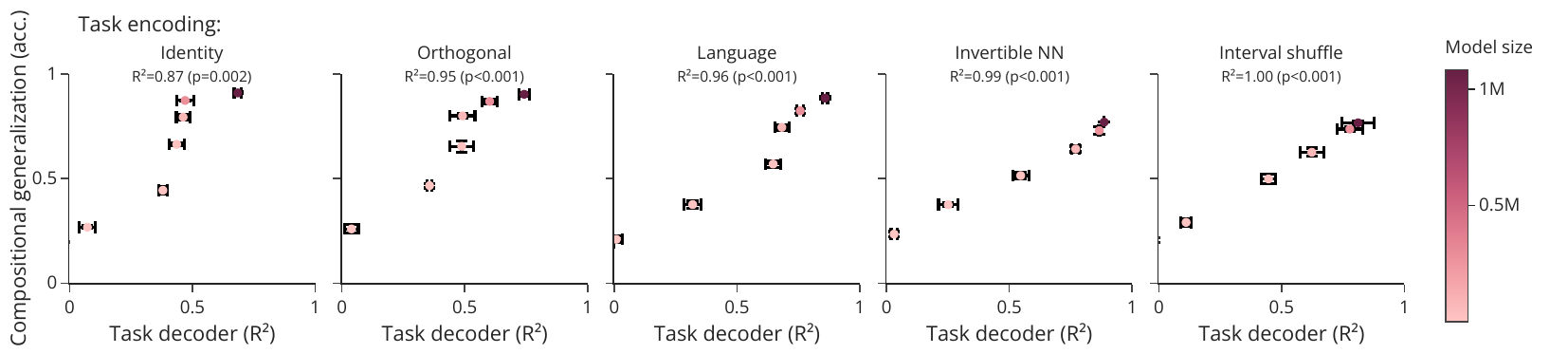}
    \caption{\textbf{Compositional generalization correlates with linear decodability of task constituents.} \textit{\textbf{Top}} Relationship between linear decodability of the task constituents and compositional generalization across instantiations of the compositional preference task family with $M=8$ preferences and varying $K$. \textit{\textbf{Bottom}} Relationship between linear decodability of the task constituents and compositional generalization across different task encodings for varying model sizes on the compositional preference task family with $M=16$, $K=3$. Error bars denote SEM over three seeds. \textit{\textbf{Top/Bottom}} We report the $R^2$ and corresponding p-value for an ordinary least square estimator in the facet titles.}
    \label{appfig:task-decoding-preference}
\end{figure}

In Figure~\ref{appfig:scaling-preference} we show, that scaling data and model size leads to compositional generalization on the compositional preference task family.
This holds for various linear and nonlinear task encodings as further shown in Table~\ref{apptab:task-encodings-preference}.
Note, that we are not able to evaluate the fewshot task encoding of the main text since the resulting input dimension is prohibitively large.
Figure~\ref{appfig:task-support-preference} shows that non-compositional and disconnected task support as well as rarely encountering a few modules interferes with compositional generalization.
And finally, Figure~\ref{appfig:task-decoding-preference} demonstrates that also on the compositional preference task family, compositional generalization and linear decodability of the task constituents from the hidden activations of the model are correlated.

\input{Tables/task-encoding_preference}

\subsection{Scaling data leads to compositional generalization in transformers}
In addition to using multilayer perceptrons as done in the main text, we investigate whether the widely used transformer architecture similarly achieves compositional generalization through scale.
For this purpose, we train a decoder-only transformer that takes a sequence consisting of the task constituents followed by the task inputs as input and predicts the corresponding labels.
Given that the transformer contains multilayer perceptrons in the feedforward blocks, we expect it to be similarly capable of compositional generalization at scale.

\begin{figure}[t]
  \centering
  \includegraphics[width=0.85\textwidth]{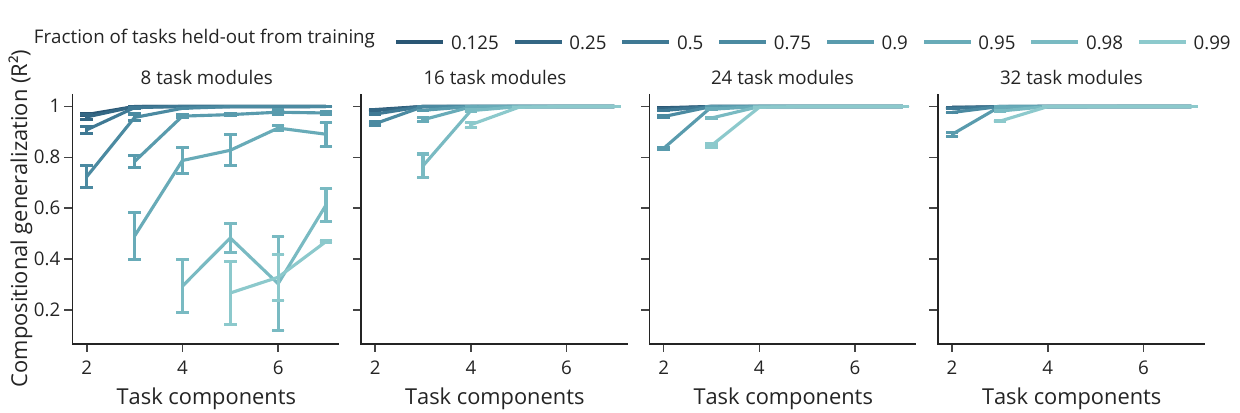}
  \includegraphics[width=0.6\textwidth]{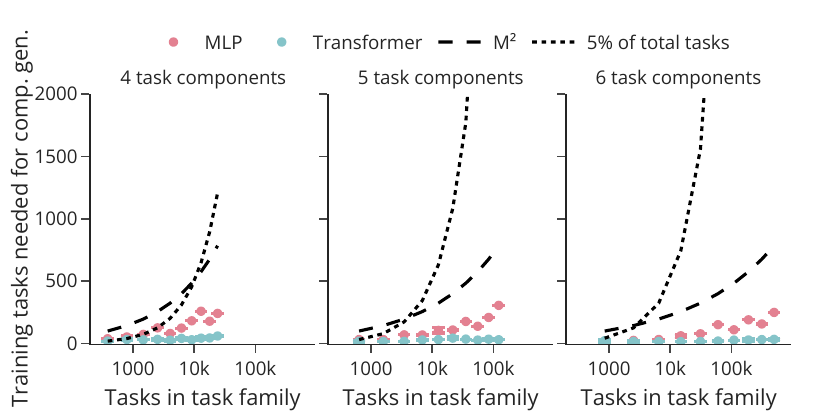}
  \caption{\textbf{Scaling data leads to compositional generalization in transformers.} \textit{\textbf{Top}} Scaling the number of training tasks by increasing the number of modules, task components or decreasing the fraction of tasks held-out from training leads to compositional generalization in transformers on the hyperteacher task family. \textit{\textbf{Bottom}} Transformers require less training tasks to achieve compositional generalization ($R^2>0.95$) compared to multilayer perceptrons (MLP).}
  \label{appfig:scaling-hyperteacher-transformer}
\end{figure}

The top of Figure~\ref{appfig:scaling-hyperteacher-transformer} confirms this, showing that scaling data similarly leads to compositional generalization in transformers.
Compared to the multilayer perceptron, the transformer requires less distinct training tasks to achieve compositional generalization, as can be observed at the bottom of Figure~\ref{appfig:scaling-hyperteacher-transformer}.

\subsection{Scaling data leads to compositional generalization in a deep hyperteacher}

\begin{figure}[t]
  \centering
  \includegraphics[width=0.6\textwidth]{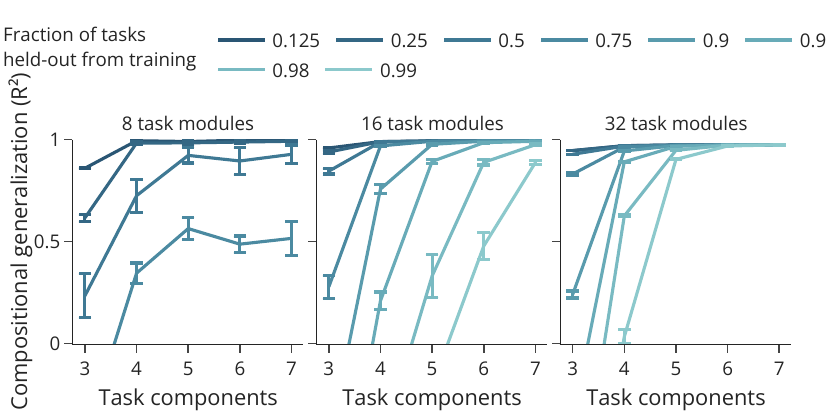}
  \caption{\textbf{Scaling data leads to compositional generalization on a deep hyperteacher.} Scaling the number of training tasks by increasing the number of modules, task components or decreasing the fraction of tasks held-out from training leads to compositional generalization on the hyperteacher task family with a deep task network with three hidden layers.}
  \label{appfig:scaling-hyperteacher-deep}
\end{figure}

In the main text, we focus on a hyperteacher with a task network with a single hidden layer.
A natural question is how increasing the difficulty of the hyperteacher by equipping the task network with multiple hidden layers affects our results.
For this purpose, we reproduce the data scaling plot shown in Figure~\ref{fig:scaling} of Section~\ref{sec:scaling} for a hyperteacher with three hidden layers, each with $16$ hidden neurons.
Figure~\ref{appfig:scaling-hyperteacher-deep} demonstrates that while this makes the task noticeably more difficult, it reproduces our finding that scaling data leads to compositional generalization.

\subsection{Difficulty of image composition task family}
\begin{figure}[t]
    \centering
    \includegraphics[width=0.49\textwidth]{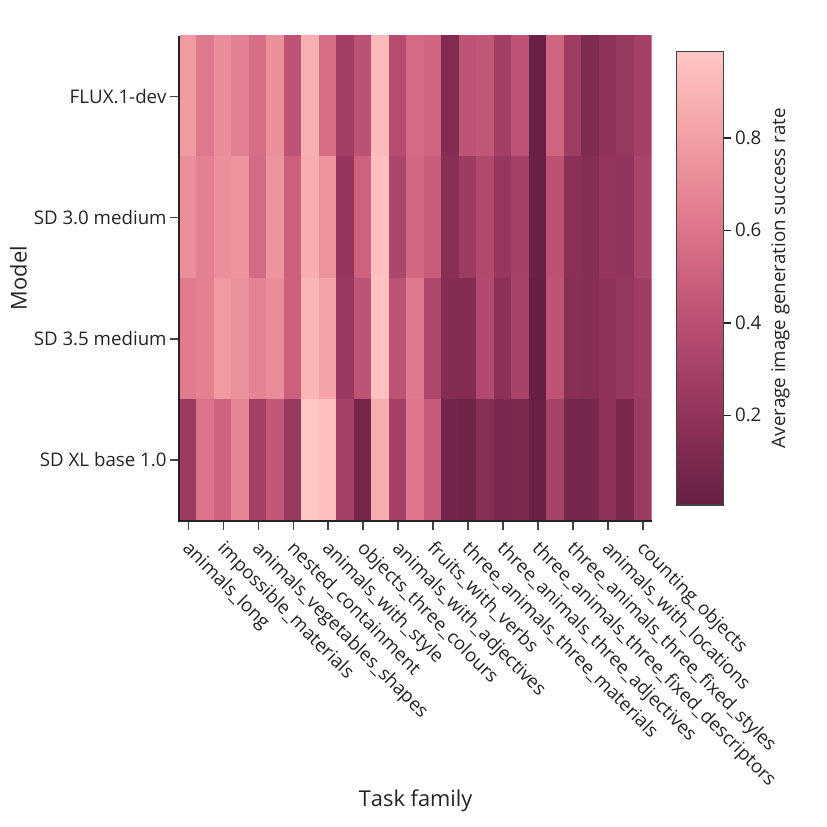}
    \includegraphics[width=0.49\textwidth]{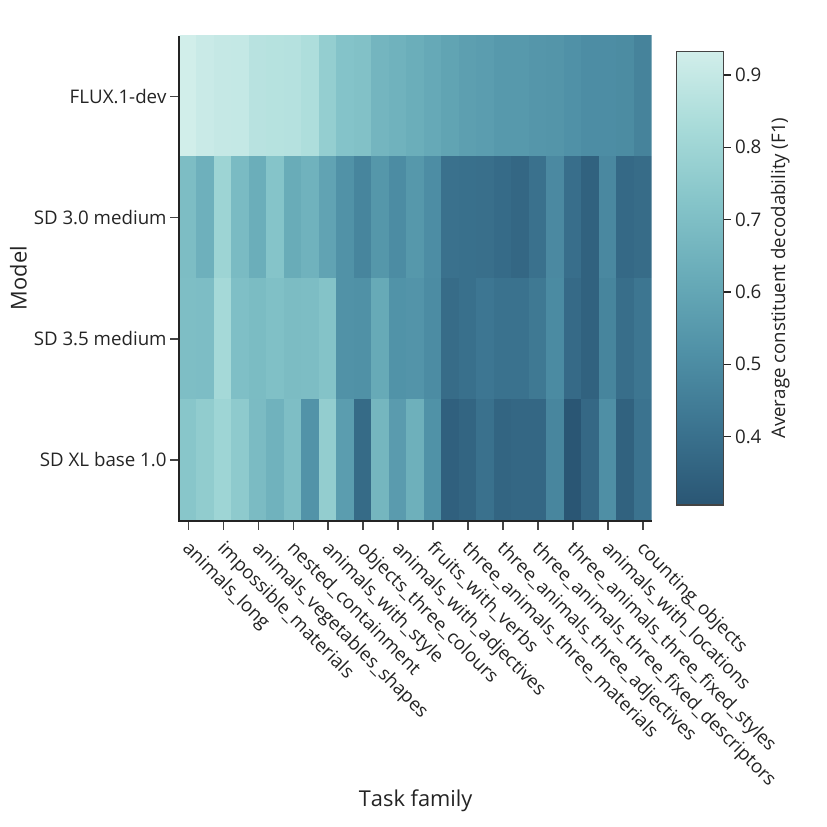}
    \caption{\textbf{Image composition task difficulties sorted by linear decodability of task constituents.} \textbf{Left} Average image generation success for each image composition task family for different text-to-image generation models. \textbf{Left} Average linear decodability of constituents 
 from hidden activations for each image composition task family for different text-to-image generation models.}
    \label{appfig:imgen-task-difficulty}
\end{figure}

To complement Figure~\ref{fig:image-generation} of the main text, Figure~\ref{appfig:imgen-task-difficulty} shows the task difficulty as well as the average linear decodability of the task constituents from the hidden activations for all the image composition task families and text-to-image generation models we consider.

\section{Experimental details}
\label{appsec:experimental-details}

\subsection{Task families}

We create the hyperteacher task family described in Section~\ref{sec:hyperteacher} with $I=16$ input neurons, $H=16$ hidden neurons and $O=16$ output neurons to create a family of compositional regression tasks to be learned by a student.
We sample teacher modules from a truncated normal distribution with zero mean and standard deviation $\sqrt{\frac{\sqrt{3}}{I}}$.
Each teacher module also has a bias term that is sampled from the uniform distribution over the interval $[0, 0.5]$.
We sample the fixed readout matrix shared by all tasks from a truncated normal distribution with zero mean and standard deviation $\sqrt{\frac{\sqrt{3}}{H}}$.
These values have been picked as to ensure that the hyperteacher creates tasks of sufficient diversity and difficulty.

For the definition of the compositional preference task family, please refer to \citep{schug_discovering_2024}.
Following the notation of the hyperteacher, we denote the number of possible preferences as $M$ and the maximum number of preferences combined into a task as $K$.
\citep{schug_discovering_2024} uses $M=8$ number of possible preferences throughout.
We increase the difficulty of the tasks by increasing the number of preference features to $M=16$ in Figure~\ref{appfig:scaling-preference} bottom, Figure~\ref{appfig:task-support-preference} and Figure~\ref{appfig:task-decoding-preference} and to $M=32$ in Figure~\ref{appfig:scaling-preference} top.

\subsection{Task encoding}

We employ the invertible neural network introduced by \citep{dinh_density_2017} with $5$ layers ($3$ for the compositional preference task family) that consists of a series of bijective transformations, to obtain a nonlinear but information-preserving encoding of the task constituents.
In addition, we use the Interval Shift function defined in Algorithm~\ref{alg:interval-shuffle} as another example of a nonlinear but bijective task encoding.

\begin{algorithm}
\caption{Interval Shift}
\label{alg:interval-shuffle}
\begin{algorithmic}[1]
\State \textbf{Number of intervals:} $N$
\Function{$\varphi$}{$z, r$}
    \State $\pi \gets \text{GeneratePermutation}(N, r)$ \Comment{Generate permutation of [0,N-1] using seed r}
    \State $i \gets \lfloor N \cdot z \rfloor$ \Comment{Determine which interval $z$ belongs to}
    \State $\alpha \gets N \cdot z - i$ \Comment{Fractional position within interval}
    \State $j \gets \pi(i)$ \Comment{Apply permutation}
    \State \Return $\frac{j + \alpha}{N}$ \Comment{Preserve relative position within new interval}
\EndFunction
\end{algorithmic}
\end{algorithm}

\subsection{Training task support}
\label{appsec:training-task-support}
We investigate the effect of various procedures to construct the training task support on compositional generalization, which we describe in the following.
For this purpose, consider the graph $G = (V, E)$ where $V = \{\bm{z} \in \mathcal{Z} \mid p^{\text{train}}(\bm{z}) > 0\}$ is the set of vertices and $E = \{(\bm{z}, \bm{z}') \in V \times V \mid \exists i \in \{1,2,\ldots,K\} \text{ such that } z_i = z'_i\}$ is the set of edges.

\begin{itemize}[leftmargin=*]
    \item \textbf{Random}: Samples a random subset of vertices until a graph that is both compositional and connected is found.
    \item \textbf{Balanced}: Similar to Random, but ensures that each module appears with equal frequency in the training distribution using a greedy search over vertices.
    \item \textbf{Non-compositional}: Holds out all tasks that contain one random but fixed module.
    \item \textbf{Disconnected}: Divides modules into two disjoint subsets and only uses vertices that use modules from either subset but do not contain modules from both subsets.
    \item \textbf{Popular modules}: 
    Defines a set of $P$ popular modules and only includes vertices that contain at least one popular module. $P$ is determined such that the fraction of held-out tasks can be satisfied as specified. If it cannot be exactly satisfied, one module that is not in the set of popular modules receives additional vertices. Additionally ensures that the resulting set is compositional and connected.
    \item \textbf{Unpopular modules}:
    Defines a set of $U$ unpopular modules and includes all vertices that do not contain any unpopular module. For each unpopular module, one vertex that includes the unpopular module and otherwise only not unpopular modules is added. $U$ is determined such that the fraction of held-out tasks can be satisfied as specified. If it cannot be exactly satisfied, one unpopular module receives additional vertices. Additionally ensures that the resulting set is compositional and connected.
\end{itemize}

\subsection{Task constituent decoding}

On the hyperteacher and compositional preference task family, we fit a linear decoder using ridge regression to predict the task constituents given the hidden activations of a particular layer of the multilayer perceptron solving the task.
For this purpose, we train on pairs of hidden activations and ground truth task constituents from the training distribution, $p^{\text{train}}(\bm{z})$, and evaluate the performance of the decoder on held-out tasks, reporting the coefficient of determination ($R^2$ score).
Throughout, we employ a regularizer of $\lambda=1.0$ for the ridge regression.

\subsection{Architecture}
Unless specified otherwise, we use a multilayer perceptron with four hidden layers with $1024$ hidden neurons each for the hyperteacher and with two hidden layers for the compositional preference task family.

The transformer in Section~\ref{appsec:additional-results} is causally masked and consists of $4$ layers with $4$ attention heads, a model dimension of $256$, a feedforward dimension of $1024$ and separate projection matrices for the task constituents, inputs and the output.

\subsection{Hyperparameters}
Throughout our experiments, we use the AdamW optimizer \citep{loshchilov_decoupled_2019} with a batch size of $128$.
On the hyperteacher task family, we use a mean-squared error loss, on the compositional preference task family, we use a cross-entropy loss.
We performed an initial grid search over the learning rate and weight decay to find a common set of hyperparameters for all experiments on the hyperteacher task family and a common set of hyperparameters for all experiments on the compositional preference tasks respectively.
We report the search grid in Table~\ref{apptab:hyperparameters}.

\begin{table}[htb]
\centering
\caption{Hyperparameters for experiments on the hyperteacher task family and compositional preference task family. Lists of values denote parameters explored via grid search with a bold number indicating the value found to perform best and used throughout the experiments.}
\label{apptab:hyperparameters}
\begin{tabular}{@{}lll@{}}
\toprule
Parameter     & Hyperteacher                          & Compositional preferences                     \\ \midrule
\texttt{learning\_rate} & $[0.001, 0.003, 0.0001, \bm{0.0003}]$ & $[0.001, 0.003, \bm{0.0001}, 0.0003]$ \\
\texttt{weight\_decay}  & $[\bm{0.003}, 0.001, 0.0003]$         & $[0.003, \bm{0.001}, 0.0003]$         \\ \bottomrule
\end{tabular}
\end{table}

\subsection{Compositional text-to-image generation}
\label{appsec:imgen}

\paragraph{Models}
We compare four open-weight text-to-image generation models: \texttt{FLUX.1-dev} \citep[][FLUX.1 dev Non-Commercial License]{black_forest_labs_black-forest-labsflux1-dev_2025}, \texttt{SD XL base 1.0} \citep[][Open RAIL++-M License]{stability_ai_stabilityaistable-diffusion-xl-base-10_2025}, \texttt{SD 3.0 medium} \citep[][Stability AI Community License]{stability_ai_stabilityaistable-diffusion-3-medium_2025} \texttt{SD 3.5 medium} \citep[][Stability AI Community License]{stability_ai_stabilityaistable-diffusion-35-medium_2025}.

\paragraph{Experimental setup}
In the following, we describe the experimental setup illustrated in Figure~\ref{fig:image-generation}.
For each image composition task family listed below, we prompt each model on all possible image combinations to generate an image using $40$ inference steps.
During this process, we collect their hidden activations to be used by the image constituent decoder (see below).
We then prompt a VLM to judge whether the image constituents have been correctly combined into a coherent image by asking it to readout the image constituents given the image and the full set of possible constituents.
We count an image generation as successful if the image constituents generated by the VLM exhaustively match the ground truth constituents.
Here, we report results using Gemini 2.0 Flash \citep{google_deepmind_gemini_2025} as the judge.

\paragraph{Image constituent decoder}
We train standard logistic regression classifiers to predict image constituents given the hidden activations of the text-to-image generation models and report their F1 scores on held-out image compositions.
Not all layers of the respective models contain this information and we need to carefully choose layers depending on the model architecture.
In particular, the image constituents can trivially be linearly decoded from early layers of the model that encode the text prompt, as well as any layers that are directly connected to the text encoding either via residual connections or via cross-attention.
For this reason, we select layers that require the model to explicitly learn to pass information about image constituents to.
In the transformer-based models \texttt{FLUX.1-dev}, \texttt{SD 3.5 medium} and \texttt{SD 3.0 medium}, we probe the hidden activation of the penultimate transformer block's MLP.
\texttt{SD XL base 1.0} contains a bottleneck block, so we probe the hidden activations at the final layer of the bottleneck.
Since the models we are considering are diffusion models that run inference over several time steps, we must specify at what time during the inference process to collect the hidden activations.
In our experiments, we run 30 inference steps and collect hidden activations at time steps 9, 15, 21, and 30, as well as the average hidden activation over all time steps.
We concatenate all hidden activations for a given model and image and use them as the input to the linear decoder.

\paragraph{Image composition task families}
\label{appsec:imgen-tasks}
We create a variety of image composition task families that each consist of a combinatorial space of objects and attributes that need to be combined into a coherent image.
Specifically, each task has the following structure with a corresponding prompt template:
\begin{itemize}[leftmargin=*]
    \item \texttt{task\_name}: \textit{Prompt template specifying how to compose \{component1\} and \{component2\}.}
    \begin{itemize}[leftmargin=*]
        \item \textbf{component1}: module1, module2, ...
        \item \textbf{component2}: module1, module2, ...
    \end{itemize}
\end{itemize}

We list all the tasks in the following:
\input{Tables/imgen-tasks-list}

\section{Additional details}

\subsection{Compute resources}
\label{appsec:compute-resources}
We used a Linux workstation with two Nvidia RTX 3090 GPUs with 24GB of memory each for development and conducted hyperparameter searches and experiments on an internal Slurm cluster using Nvidia RTX 4090 GPUs and Nvidia A100 GPUs.

A single run of a hyperteacher experiment takes 4-10 minutes on a RTX 4090 depending on model size, a single run of a compositional preference experiment takes 50-100 minutes.
In total, reproducing all experiments reported for the hyperteacher task family with around 1000 distinct runs takes about 7 GPU days and reproducing all corresponding experiments for the compositional preference task family takes about 70 GPU days.

Generating the images for one of the image composition task families takes 8 GPU hours on an A100 for the \texttt{Flux.1-dev} model and 4 GPU hours on an RTX 4090 for each of the \texttt{SD} models.
Running all 27 tasks for all models takes a total of 23 GPU days.

\subsection{Software and libraries}
\label{appsec:software}
For the results obtained in this paper we built on free and open-source software.
We implemented our experiments in Python using JAX \citep[][Apache License 2.0]{bradbury_jax_2018}, Flax \citep[][Apache License 2.0]{heek_flax_2023}, the Deepmind Jax Ecosystem \citep[][Apache License 2.0]{babuschkin_deepmind_2020}, PyTorch \citep[BSD-style license][]{paszke_pytorch_2019}, LLM \citep[][Apache License 2.0]{willison_llm_2023} and Scikit-learn \citep[][BSD 3-Clause License]{pedregosa_scikit-learn_2011}.
We utilized WandB \citep[][MIT license]{biewald_experiment_2020} to monitor the progress and results of experiments, and Plotly  \citep[][MIT license]{inc_collaborative_2015} for generating the plots.
We use uv for Python project dependency management \citep[][MIT License]{marsh_uv_2024}.

%% file: Tables/task-encoding_preference.tex
\begin{table}
\centering
\caption{\textbf{Compositional generalization emerges across task encodings.} Comparison of the decodability of the task constituents from the hidden activations (Task decoder) and compositional generalization performance for linear and nonlinear task encodings on the compositional preference task family with $M=16$, $K=3$. For the linear Identity, Orthogonal and Language task encoding, we report the decodability of task constituents from the first layer whereas for the nonlinear Invertible NN (with 2 layers) and Interval shuffle encodings we report it for the second layer. As opposed to linear task encodings, for nonlinear task encodings the decodability of task constituents is higher in the second layer compared to the first layer suggesting that the first layer is used to linearize the task constituents. We additionally show the linear decodability of the task constituents directly from the task encoding itself (Input decoder), which allows to distinguish linear from nonlinear task encodings. $\pm$ SEM over three seeds.}
\label{apptab:task-encodings-preference}
\begin{tabular}{lrrr}
\toprule
Task encoding & Task decoder (R²) & Embedding decoder (R²) & Comp. gen. (acc.) \\
\midrule
Identity & $0.95 \pm 0.012$ & $1.00 \pm 0.000$ & $0.92 \pm 0.004$ \\
Orthogonal & $0.96 \pm 0.008$ & $1.00 \pm 0.000$ & $0.92 \pm 0.006$ \\
Language & $0.98 \pm 0.004$ & $1.00 \pm 0.000$ & $0.92 \pm 0.004$ \\
Invertible NN & $0.93 \pm 0.005$ & $0.74 \pm 0.018$ & $0.82 \pm 0.009$ \\
Interval shuffle & $0.90 \pm 0.022$ & $0.72 \pm 0.083$ & $0.84 \pm 0.017$ \\
\bottomrule
\end{tabular}
\end{table}

%% file: Tables/imgen-tasks-list.tex
\begin{itemize}[leftmargin=*]
    \item \texttt{animals\_with\_colours}: \textit{A \{colour\} \{animal\}}
    \begin{itemize}[leftmargin=*]
        \item \textbf{animal}: lion, elephant, giraffe, crocodile, bear, snake, eagle, cow, zebra, tiger
        \item \textbf{colour}: red, blue, green, yellow, orange, purple, pink, brown, black, white
    \end{itemize}
    \item \texttt{animals\_with\_style}: \textit{A \{animal\} illustrated in \{style\} style}
    \begin{itemize}[leftmargin=*]
        \item \textbf{animal}: lion, elephant, giraffe, crocodile, bear, snake, eagle, cow, zebra, tiger
        \item \textbf{style}: watercolor, pixel art, oil painting, sketch, cartoon, origami, stained glass, pop art, charcoal, clay sculpture
    \end{itemize}
    \item \texttt{animals\_with\_locations}: \textit{A \{animal\} in the \{location\}}
    \begin{itemize}[leftmargin=*]
        \item \textbf{animal}: lion, elephant, giraffe, crocodile, bear, snake, eagle, cow, zebra, tiger
        \item \textbf{location}: top left, top center, top right, middle left, center, middle right, bottom left, bottom center, bottom right
    \end{itemize}
    \item \texttt{animals\_with\_descriptor}: \textit{A \{descriptor\} \{animal\}}
    \begin{itemize}[leftmargin=*]
        \item \textbf{animal}: lion, elephant, giraffe, crocodile, bear, snake, eagle, cow, zebra, tiger
        \item \textbf{descriptor}: furry, scaly, feathered, leathery, smooth, wrinkled, young, old, three-legged, spotted
    \end{itemize}
    \item \texttt{animals\_with\_adjectives}: \textit{A image of a \{adjective\} \{animal\}}
    \begin{itemize}[leftmargin=*]
        \item \textbf{adjective}: happy, sad, angry, sleepy, curious, playful, scared, proud, surprised, bored
        \item \textbf{animal}: lion, elephant, giraffe, crocodile, bear, snake, eagle, cow, zebra, tiger
    \end{itemize}
    \item \texttt{fruits\_with\_verbs}: \textit{A \{fruit\} is \{verb\}}
    \begin{itemize}[leftmargin=*]
        \item \textbf{fruit}: apple, banana, orange, grape, strawberry, watermelon, pineapple, mango, blueberry, peach
        \item \textbf{verb}: dancing, flying, bouncing, sleeping, swimming, rolling, climbing, stretching, spinning, hiding
    \end{itemize}
    \item \texttt{counting\_animals}: \textit{An image with exactly \{number\} \{animal\}}
    \begin{itemize}[leftmargin=*]
        \item \textbf{number}: one, two, three, four, five, six, seven, eight, nine, ten
        \item \textbf{animal}: lions, elephants, giraffes, crocodiles, bears, snakes, eagles, cows, zebras, tigers
    \end{itemize}
    \item \texttt{animals\_with\_verbs}: \textit{A \{animal\} is \{verb\}}
    \begin{itemize}[leftmargin=*]
        \item \textbf{animal}: lion, elephant, giraffe, crocodile, bear, snake, eagle, cow, zebra, tiger
        \item \textbf{verb}: eating, sleeping, running, jumping, flying, swimming, climbing, dancing, playing, hiding
    \end{itemize}
    \item \texttt{counting\_objects}: \textit{An image with exactly \{number\} \{object\}}
    \begin{itemize}[leftmargin=*]
        \item \textbf{number}: one, two, three, four, five, six, seven, eight, nine, ten, eleven, twelve, thirteen, fourteen, fifteen, sixteen, seventeen, eighteen, nineteen, twenty
        \item \textbf{object}: tomatoes, onions, oranges, wolves, bears, apples, bananas, carrots, cucumbers, strawberries, lemons, cherries, grapes, peaches, pears, foxes, rabbits, cats, dogs, sheep
    \end{itemize}
    \item \texttt{nested\_containment\_animals}: \textit{A \{animal\_outer\} containing a \{animal\_middle\} containing an \{animal\_inner\}}
    \begin{itemize}[leftmargin=*]
        \item \textbf{animal\_outer}: lion, elephant, giraffe, crocodile, bear, snake, eagle, cow, zebra, tiger
        \item \textbf{animal\_middle}: lion, elephant, giraffe, crocodile, bear, snake, eagle, cow, zebra, tiger
        \item \textbf{animal\_inner}: lion, elephant, giraffe, crocodile, bear, snake, eagle, cow, zebra, tiger
    \end{itemize}
    \item \texttt{three\_animals\_three\_materials}: \textit{\{animal\_fire\} made of fire, \{animal\_ice\} made of ice, and \{animal\_wood\} made of wood}
    \begin{itemize}[leftmargin=*]
        \item \textbf{animal\_fire}: lion, elephant, giraffe, crocodile, bear, snake, eagle, cow, zebra, tiger
        \item \textbf{animal\_ice}: lion, elephant, giraffe, crocodile, bear, snake, eagle, cow, zebra, tiger
        \item \textbf{animal\_wood}: lion, elephant, giraffe, crocodile, bear, snake, eagle, cow, zebra, tiger
    \end{itemize}
    \item \texttt{three\_animals\_three\_verbs}: \textit{\{animal\_singing\} singing, \{animal\_eating\} eating, and \{animal\_sleeping\} sleeping}
    \begin{itemize}[leftmargin=*]
        \item \textbf{animal\_singing}: lion, elephant, giraffe, crocodile, bear, snake, eagle, cow, zebra, tiger
        \item \textbf{animal\_eating}: lion, elephant, giraffe, crocodile, bear, snake, eagle, cow, zebra, tiger
        \item \textbf{animal\_sleeping}: lion, elephant, giraffe, crocodile, bear, snake, eagle, cow, zebra, tiger
    \end{itemize}
    \item \texttt{three\_animals\_three\_adjectives}: \textit{A sad \{animal\_happy\}, a happy \{animal\_sad\}, and an angry \{animal\_angry\}}
    \begin{itemize}[leftmargin=*]
        \item \textbf{animal\_happy}: lion, elephant, giraffe, crocodile, bear, snake, eagle, cow, zebra, tiger
        \item \textbf{animal\_sad}: lion, elephant, giraffe, crocodile, bear, snake, eagle, cow, zebra, tiger
        \item \textbf{animal\_angry}: lion, elephant, giraffe, crocodile, bear, snake, eagle, cow, zebra, tiger
    \end{itemize}
    \item \texttt{animals\_with\_clothes}: \textit{\{animal\} wearing \{clothing\}}
    \begin{itemize}[leftmargin=*]
        \item \textbf{animal}: cat, dog, bear, lion, elephant, giraffe, monkey, zebra, tiger, panda
        \item \textbf{clothing}: hat, pair of sunglasses, scarf, bowtie, jacket, crown, tie, cape, sweater, necklace
    \end{itemize}
    \item \texttt{animals\_with\_clothes\_and\_food}: \textit{\{animal\} wearing \{clothing\} eating \{food\}}
    \begin{itemize}[leftmargin=*]
        \item \textbf{animal}: cat, dog, bear, lion, elephant, giraffe, monkey, zebra, tiger, panda
        \item \textbf{clothing}: hat, pair of sunglasses, scarf, bowtie, jacket, crown, tie, cape, pair of pants, pair of boots
        \item \textbf{food}: pizza, banana, ice cream, cake, hamburger, apple, watermelon, donut, sandwich, salad
    \end{itemize}
    \item \texttt{animals\_with\_food\_eyes\_and\_clothes}: \textit{A \{animal\} with \{food\} as eyes wearing \{clothing\}}
    \begin{itemize}[leftmargin=*]
        \item \textbf{animal}: cat, bear, lion, elephant, giraffe, monkey, zebra, panda, wolf, rabbit
        \item \textbf{food}: strawberries, oranges, burgers, watermelons, donuts, cookies, cupcakes, pizza, lemons, tomatoes
        \item \textbf{clothing}: crown, cowboy hat, scarf, cape, hawaiian shirt, leather jacket, pair of pants, tuxedo, raincoat, sunglasses
    \end{itemize}
    \item \texttt{stacked\_foods}: \textit{\{food\_top\} on top of \{food\_middle\} on top of \{food\_bottom\}}
    \begin{itemize}[leftmargin=*]
        \item \textbf{food\_top}: burger, pizza, salad, Sushi, taco, donut, Ice cream, pancake, Spaghetti, sandwich
        \item \textbf{food\_middle}: burger, pizza, salad, sushi, taco, donut, ice cream, pancake, spaghetti, sandwich
        \item \textbf{food\_bottom}: burger, pizza, salad, sushi, taco, donut, ice cream, pancake, spaghetti, sandwich
    \end{itemize}
    \item \texttt{stacked\_animals}: \textit{\{animal\_top\} on top of \{animal\_middle\} on top of \{animal\_bottom\}}
    \begin{itemize}[leftmargin=*]
        \item \textbf{animal\_top}: lion, elephant, giraffe, crocodile, bear, snake, eagle, cow, zebra, tiger
        \item \textbf{animal\_middle}: lion, elephant, giraffe, crocodile, bear, snake, eagle, cow, zebra, tiger
        \item \textbf{animal\_bottom}: lion, elephant, giraffe, crocodile, bear, snake, eagle, cow, zebra, tiger
    \end{itemize}
    \item \texttt{animals\_chasing\_chain}: \textit{\{animal1\} chasing \{animal2\} chasing \{animal3\}}
    \begin{itemize}[leftmargin=*]
        \item \textbf{animal1}: lion, elephant, giraffe, crocodile, bear, snake, eagle, cow, zebra, tiger
        \item \textbf{animal2}: lion, elephant, giraffe, crocodile, bear, snake, eagle, cow, zebra, tiger
        \item \textbf{animal3}: lion, elephant, giraffe, crocodile, bear, snake, eagle, cow, zebra, tiger
    \end{itemize}
    \item \texttt{nested\_containment}: \textit{A \{container1\} containing a \{container2\} containing a \{object\}}
    \begin{itemize}[leftmargin=*]
        \item \textbf{container1}: transparent cube, glass jar, wooden box, metal safe, woven basket, leather bag, ceramic pot, copper kettle, crystal sphere, rubber ball
        \item \textbf{container2}: small treasure chest, porcelain teacup, silk pouch, stone bowl, paper envelope, cardboard tube, tin can, shell, gold locket, velvet case
        \item \textbf{object}: diamond, living butterfly, ticking clock, miniature planet, flickering flame, drop of mercury, hologram, glowing ember, snowflake, single cell organism
    \end{itemize}
    \item \texttt{impossible\_materials}: \textit{A \{object\} made entirely of \{material\} sitting on a \{surface\}}
    \begin{itemize}[leftmargin=*]
        \item \textbf{object}: chair, bicycle, bookshelf, piano, computer, refrigerator, watch, umbrella, camera, guitar
        \item \textbf{material}: liquid water, fire, smoke, mirrors, ice, tree bark, glass noodles, gelatin, paper, soap bubbles
        \item \textbf{surface}: clouds, ocean waves, melting ice, sand dunes, moss, broken glass, spiderwebs, lily pads, autumn leaves, foam
    \end{itemize}
    \item \texttt{three\_animals\_three\_fixed\_styles}: \textit{A \{animal\_pixel\} in pixel art, a \{animal\_oil\} in oil painting, and a \{animal\_cartoon\} in cartoon style}
    \begin{itemize}[leftmargin=*]
        \item \textbf{animal\_pixel}: lion, elephant, giraffe, crocodile, bear, snake, eagle, cow, zebra, tiger
        \item \textbf{animal\_oil}: lion, elephant, giraffe, crocodile, bear, snake, eagle, cow, zebra, tiger
        \item \textbf{animal\_cartoon}: lion, elephant, giraffe, crocodile, bear, snake, eagle, cow, zebra, tiger
    \end{itemize}
    \item \texttt{three\_animals\_three\_fixed\_descriptors}: \textit{A \{animal\_furry\} with fur, a \{animal\_scaly\} with scales, and a \{animal\_feathered\} with feathers}
    \begin{itemize}[leftmargin=*]
        \item \textbf{animal\_furry}: lion, elephant, giraffe, crocodile, bear, snake, eagle, cow, zebra, tiger
        \item \textbf{animal\_scaly}: lion, elephant, giraffe, crocodile, bear, snake, eagle, cow, zebra, tiger
        \item \textbf{animal\_feathered}: lion, elephant, giraffe, crocodile, bear, snake, eagle, cow, zebra, tiger
    \end{itemize}
    \item \texttt{animals\_long}: \textit{An image with \{one \{animal\}\}$^{1,2,3}$}
    \begin{itemize}[leftmargin=*]
        \item \textbf{animals}: giraffe, lion, elephant, crocodile, bear, snake, eagle, cow, zebra, tiger, rhino, hippo, wolf, fox, deer, monkey, panda, koala, kangaroo, penguin
    \end{itemize}
    \item \texttt{objects\_three\_colours}: \textit{An image with \{one \{object\}\}$^{1,2,3}$}
    \begin{itemize}[leftmargin=*]
        \item \textbf{objects}: red cube, green cube, blue cube, blue sphere, orange sphere, red sphere, green cylinder, purple cylinder, yellow cylinder, yellow cone, blue cone, green cone, purple pyramid, red pyramid, orange pyramid, orange cuboid, yellow cuboid, blue cuboid
    \end{itemize}
    \item \texttt{fruits\_veggies\_long}: \textit{An image with \{one \{fruit\_veggie\}\}$^{1,2,3}$}
    \begin{itemize}[leftmargin=*]
        \item \textbf{fruits\_veggies}: apple, banana, orange, grape, strawberry, watermelon, pineapple, mango, blueberry, peach, lemon, cherry, carrot, broccoli, tomato, cucumber, potato, onion, pepper, lettuce
    \end{itemize}
    \item \texttt{animals\_vegetables\_shapes}: \textit{An image with \{one \{element\}\}$^{1,2,3}$}
    \begin{itemize}[leftmargin=*]
        \item \textbf{animals\_vegetables\_shapes}: lion, elephant, giraffe, tiger, bear, zebra, monkey, carrot, broccoli, potato, tomato, cucumber, onion, pepper, cube, sphere, cylinder, cone, pyramid, triangular prism
    \end{itemize}
\end{itemize}